\documentclass{article}

\usepackage[preprint]{neurips_2023}

\usepackage[utf8]{inputenc} 
\usepackage[T1]{fontenc}    
\usepackage{hyperref}       
\usepackage{url}            
\usepackage{booktabs}       
\usepackage{amsfonts}       
\usepackage{nicefrac}       
\usepackage{microtype}      
\usepackage{xcolor}         

\usepackage{enumitem} 
\usepackage{xspace}
\usepackage{graphicx}
\usepackage{booktabs}
\usepackage{multirow}
\usepackage{colortbl}
\usepackage{amsmath}
\usepackage[capitalize]{cleveref}
\usepackage{natbib}
\usepackage{wrapfig}

\usepackage[most]{tcolorbox}  
\usepackage{caption}

\definecolor{Gray}{gray}{0.9}
\definecolor{mylightblue}{rgb}{0.8, 0.9, 1.0}

\definecolor{myblue}{HTML}{2E5AA8}
\hypersetup{
  colorlinks,
  linkcolor=myblue,
  citecolor=myblue,
  urlcolor=myblue
}

\setcitestyle{numbers}
\setcitestyle{square}
\setcitestyle{comma}

\newcommand{\dataname}{\textsc{DocQA-RL-1.6K}}
\newcommand{{\modelname}}{\textsc{QwenLong-L1}}

\title{{\modelname}: Towards Long-Context Large Reasoning Models with Reinforcement Learning}

\author{Fanqi Wan, Weizhou Shen, Shengyi Liao, Yingcheng Shi, Chenliang Li, \\ \bf{Ziyi Yang, Ji Zhang, Fei Huang, Jingren Zhou, Ming Yan\thanks{$\;$Corresponding author.}} \\
Tongyi Lab, Alibaba Group
}

\def\huggingface{\raisebox{-1.5pt}{\includegraphics[height=1.05em]{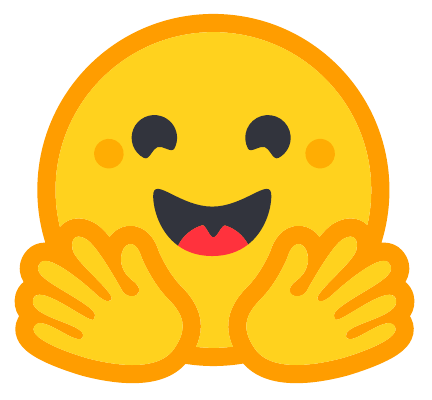}}}
\def\modelscope{\raisebox{-1.5pt}{\includegraphics[height=0.85em]{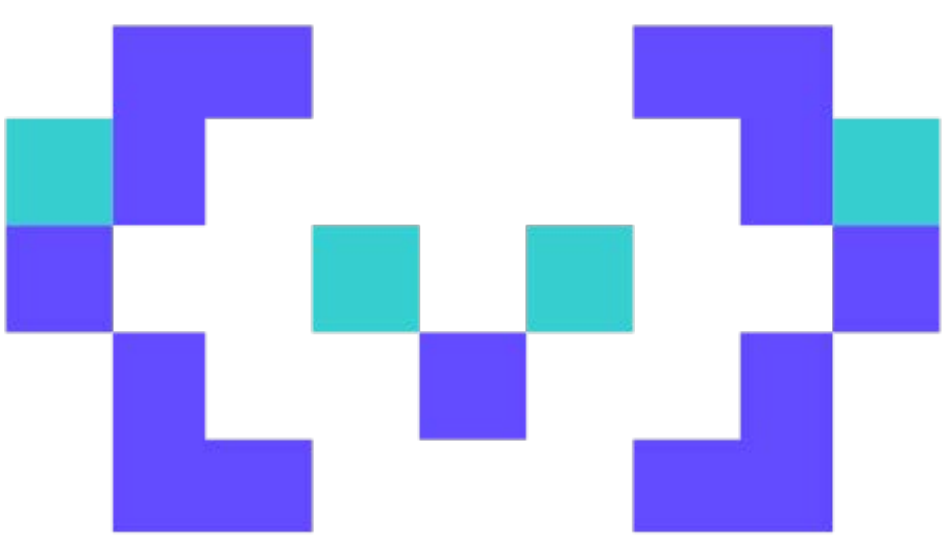}}}
\def\github{\raisebox{-1.5pt}{\includegraphics[height=1.05em]{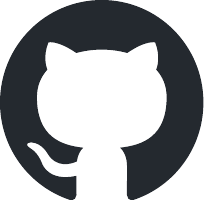}}}

\newcommand{\ghlink}{https://github.com/Tongyi-Zhiwen/QwenLong-L1}
\newcommand{\hflink}{https://huggingface.co/Tongyi-Zhiwen/QwenLong-L1-32B}
\newcommand{\mslink}{https://modelscope.cn/models/iic/QwenLong-L1-32B}

\begin{document}

\maketitle
\begin{center}
\vspace{-0.8cm}
\begin{tabular}{rc}
\github & \url{\ghlink}\\
\huggingface & \url{\hflink}\\
\modelscope & \url{\mslink}\\
\end{tabular}
\vspace{0.2cm}
\end{center}

\begin{abstract}
Recent large reasoning models (LRMs) have demonstrated strong reasoning capabilities through reinforcement learning (RL). These improvements have primarily been observed within the short-context reasoning tasks. In contrast, extending LRMs to effectively process and reason on long-context inputs via RL remains a critical unsolved challenge. To bridge this gap, we first formalize the paradigm of \textit{long-context reasoning RL}, and identify key challenges in \textit{suboptimal training efficiency} and \textit{unstable optimization process}. To address these issues, we propose {\modelname}, a framework that adapts short-context LRMs to long-context scenarios via \textit{progressive context scaling}. Specifically, we utilize a warm-up supervised fine-tuning (SFT) stage to establish a robust initial policy, followed by a curriculum-guided phased RL technique to stabilize the policy evolution, and enhanced with a difficulty-aware retrospective sampling strategy to incentivize the policy exploration. Experiments on seven long-context document question-answering benchmarks demonstrate that {\modelname}-32B outperforms flagship LRMs like OpenAI-o3-mini and Qwen3-235B-A22B, achieving performance on par with Claude-3.7-Sonnet-Thinking, demonstrating leading performance among state-of-the-art LRMs. This work advances the development of practical long-context LRMs capable of robust reasoning across information-intensive environments.

\end{abstract}

\begin{figure}[!ht]
    \centering
    \vspace{-0.2cm}
    \includegraphics[width=0.999\textwidth]{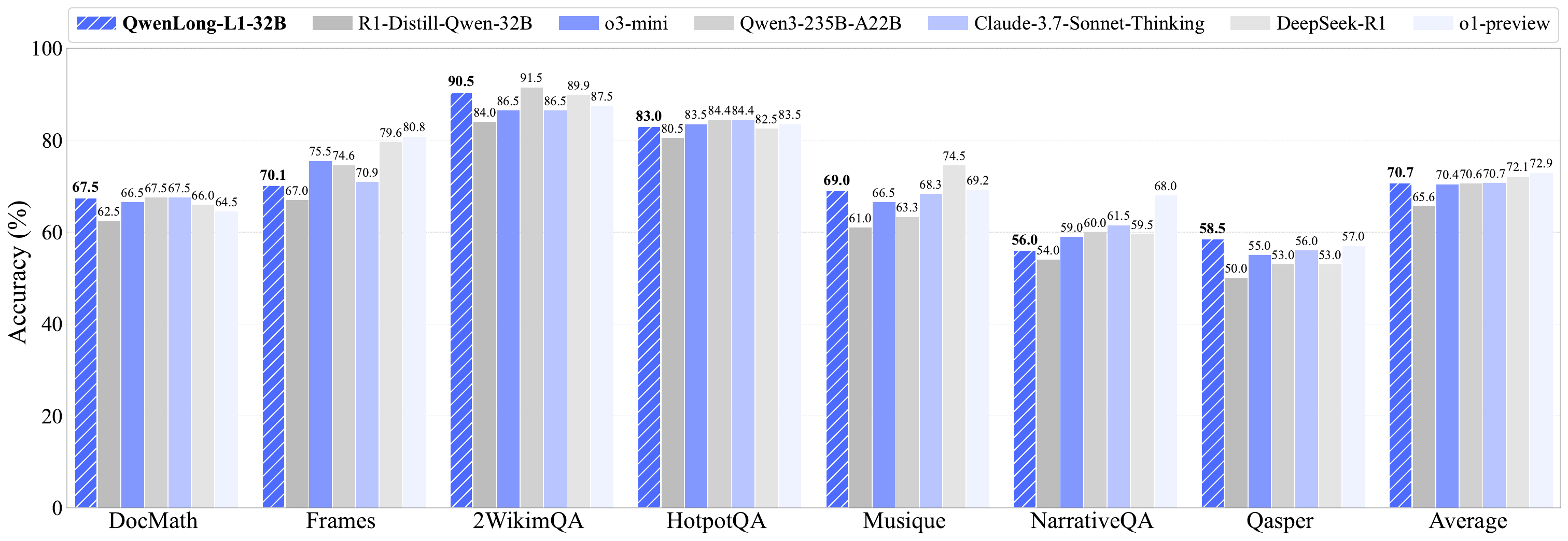}
 \caption{Overall results of {\modelname} across seven long-context reasoning benchmarks. Starting from R1-Distill-Qwen-32B, {\modelname}-32B achieves an average gain of 5.1 points, surpassing OpenAI-o3-mini, Qwen3-235B-A22B, and comparable to Claude-3.7-Sonnet-Thinking.}
\label{fig:overall_results}
\end{figure}

\clearpage

\section{Introduction}

\begin{figure}[!t]
    \centering
    \includegraphics[width=0.999\textwidth]{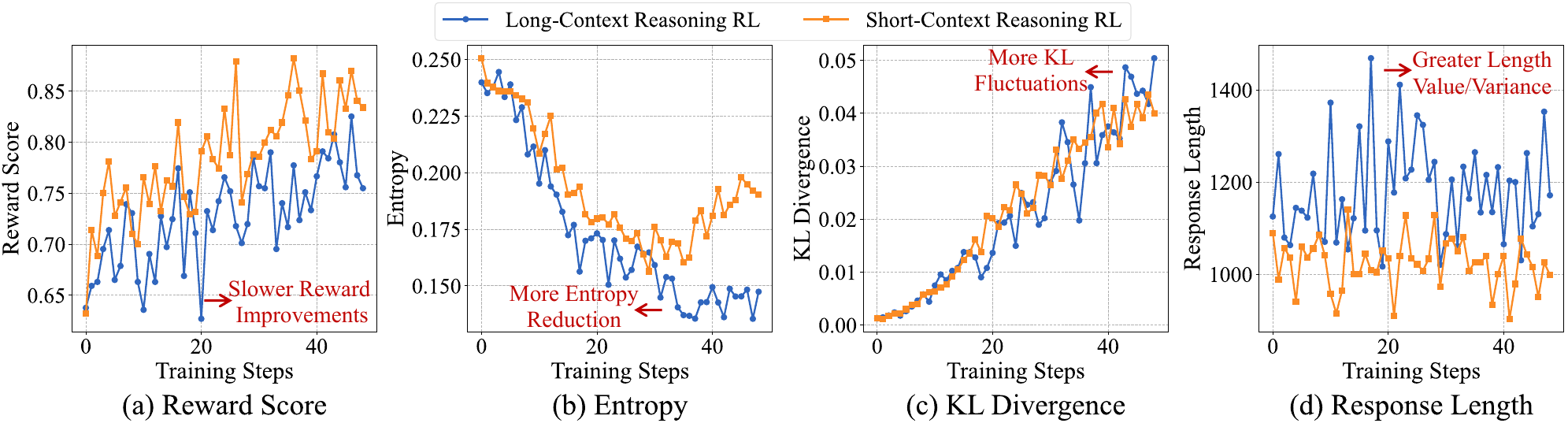}
 \caption{Comparison of training dynamics between short-context and long-context reasoning RL. The long-context reasoning RL demonstrates two key challenges: \textit{suboptimal training efficiency}, with slower improvements in reward score caused by more reduction in entropy, and \textit{unstable optimization process}, with more fluctuations in KL divergence introduced from greater variance in longer output.}
\label{fig:training_line}
\vspace{-0.3cm}
\end{figure}

Recent breakthroughs in large reasoning models (LRMs) have showcased significant improvements in reasoning capabilities, achieving performance comparable to human experts in complex problem-solving scenarios~\citep{xu2025towards}. These advancements, exemplified by OpenAI-o1~\citep{2409_openai_o1,openai2024openaio1card}, DeepSeek-R1~\citep{deepseek-r1-preview,guo2025deepseek}, and Qwen-QwQ~\citep{qwq-32b-preview,qwq-32b}, have sparked extensive research efforts to explore and enhance a broad spectrum of reasoning tasks through reinforcement learning (RL), ranging from foundational logical reasoning~\citep{tinyzero,xie2025logic} to advanced challenges in programming~\citep{openr1,el2025competitive} and mathematics~\citep{deepscaler2025,sky-t1-7b}, with innovations in RL algorithms driving progress in reasoning quality enhancements~\citep{cui2025process,yuan2025s,Liu2025UnderstandingRT,yu2025dapo}.

Following RL fine-tuning, LRMs exhibit a phenomenon analogous to human ``slow thinking''~\citep{daniel2017thinking}, characterized by the emergence of sophisticated problem-solving strategies such as divide-and-conquer and backtracking mechanisms in their extended chain-of-thought (CoT) reasoning outputs~\citep{wu2024comparative}. While this process enhances reasoning performance on short context tasks (e.g., 4K tokens)~\citep{wen2025light,liu2025comprehensivesurveylongcontext}, its scalability to long-context scenarios (e.g., 120K tokens), which requires robust contextual grounding and multi-step reasoning, remains unexplored. This limitation poses a significant barrier to practical applications requiring interaction with external knowledge, such as deep research~\citep{google2024deepresearch,openai2025deepresearch,perplexity2025deepresearch}, where LRMs must collect and process information from knowledge-intensive environments.

To shed light on this topic, we first introduce the concept of \textit{long-context reasoning RL}. Different from \textit{short-context reasoning RL}, which primarily relies on internal knowledge stored within model parameters, long-context reasoning RL instead necessitates that LRMs perform retrieval and grounding of relevant information from long-context inputs, followed by generation of reasoning chains based on the incorporated information~\citep{guu2020retrieval,ram2023context,yao2023react}. To illustrate the differences between short-context and long-context reasoning RL, we conduct a preliminary experiment to compare the training dynamics in Figure \ref{fig:training_line}. Our results reveal that long-context reasoning RL exhibits \textit{suboptimal training efficiency} compared to the short-context counterpart with \textit{(a) delayed reward convergence}. This discrepancy stems from \textit{(b) marked reduction in output entropy} when processing long-context inputs, which restricts exploratory behavior during policy optimization. Furthermore, we identify \textit{unstable optimization process}, evidenced by \textit{(c) intermittent spikes in KL divergence}. These instabilities are introduced by the inherent variance amplification due to \textit{(d) longer output length with heterogeneous input length distributions}, leading to greater variability during policy updating.

\begin{figure}[!t]
    \centering
    \includegraphics[width=0.999\textwidth]{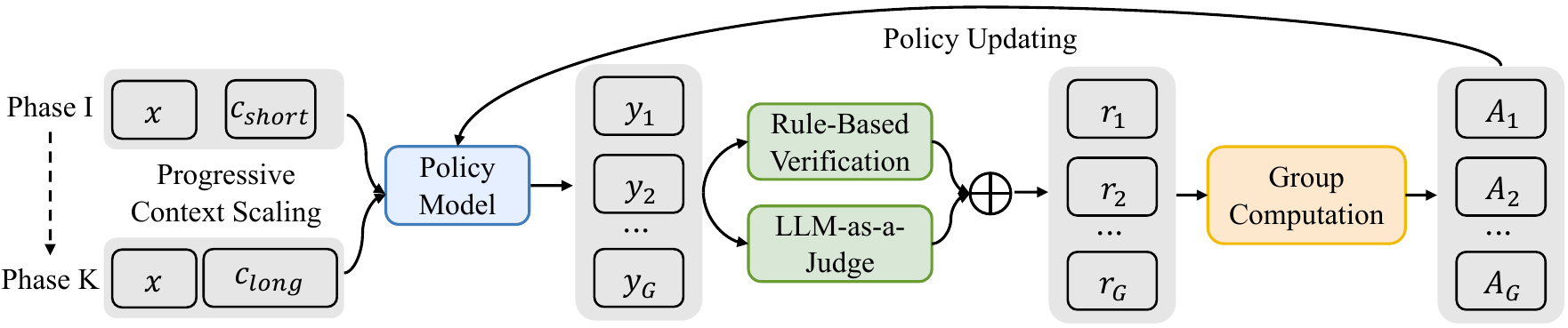}
 \caption{Overview of {\modelname}, which is a novel \textit{long-context reasoning RL} training framework. The proposed framework integrates group-relative RL algorithms, hybrid reward mechanisms, and progressive context scaling strategies to enable stable adaptation from short-context to long-context LRMs with robust contextual grounding and multi-step reasoning capabilities.}
\label{fig:method}
\vspace{-0.3cm}
\end{figure}

To address these challenges, we propose {\modelname}, a novel RL framework designed to facilitate the transition of LRMs from short-context proficiency to robust long-context generalization, as shown in Figure \ref{fig:method}. Inspired by recent studies on context extension during pretraining~\citep{fu2024data,xiong2024effective,gao2024train}, our framework enhances short-context LRMs through progressive context scaling during RL training. The framework comprises three core components: a warm-up supervised fine-tuning (SFT) phase to initialize a robust policy, a curriculum-guided RL phase that facilitates stable adaptation from short to long contexts, and a difficulty-aware retrospective sampling mechanism that adjusts training complexity across stages to incentivize policy exploration. Leveraging recent RL algorithms, including GRPO \citep{shao2024deepseekmath} and DAPO \citep{yu2025dapo}, our framework integrates hybrid reward functions combining rule-based and model-based binary outcome rewards to balance precision and recall. Through strategic utilization of group relative advantages during policy optimization, it guides LRMs to learn effective reasoning patterns essential for long-context reasoning scenarios, resulting in robust long-context grounding and superior reasoning capabilities. 

In our experiments, we focus on document question answering (DocQA)~\citep{yang2018hotpotqa,kovcisky2018narrativeqa,ho2020constructing} as a representative real-world long-context reasoning task. Specifically, we introduce {\dataname}, a specialized RL training dataset comprising 1.6K DocQA problems spanning mathematical, logical, and multi-hop reasoning domains. Experimental results across seven long-context DocQA benchmarks demonstrate the superiority of {\modelname} compared to various proprietary and open-source LRMs. Notably, {\modelname}-14B achieves superior performance over Gemini-2.0-Flash-Thinking and Qwen3-32B, while {\modelname}-32B outperforms OpenAI-o3-mini, Qwen3-235B-A22B, and even matches Claude-3.7-Sonnet-Thinking. Our analysis further identifies several critical insights in long-context reasoning RL optimization: (1) progressive context scaling promotes higher entropy and stabilizes KL divergence, enhancing training efficiency; (2) SFT proves to be an economical way for performance enhancement, whereas RL unlocks the potential to achieve optimal performance; (3) RL naturally fosters specialized long-context reasoning behaviors that boost final performance, but imitating these behaviors do not translate into gains when applied to SFT. Our key contributions are summarized as follows:

\begin{itemize}[leftmargin=1em]

\item We conceptualize the paradigm of \textit{long-context reasoning RL} and identify its unique challenges, making a further step towards developing practical long-context LRMs capable of grounding and integrating information for complex, real-world reasoning scenarios.

\item We present {\modelname}, the first RL framework designed for long-context LRMs. Through progressive context scaling, {\modelname} enables stable short-to-long context adaptation via group-relative RL optimization and hybrid reward mechanisms.

\item We showcase the effectiveness of {\modelname} through comprehensive experiments across seven long-context document question answering benchmarks. Our results reveal that {\modelname} achieves substantial performance gains compared to cutting-edge LRMs, offering a fundamental recipe and practice for long-context reasoning optimization.

\end{itemize}

\section{Long-Context Reasoning Reinforcement Learning}
\label{sec:method}

While existing works have explored reinforcement learning (RL) for enhancing short-context reasoning tasks~\citep{OpenReasonerZero2025,zeng2025simplerlzooinvestigatingtamingzero,jin2025search}, the extension of long-context reasoning RL remains an unresolved challenge. To this end, we introduce {\modelname}, the first RL framework specifically designed for long-context reasoning. In this section, we first describe the preliminaries, and then detail the implementation, including the RL algorithms for long-context reasoning, the progressive context scaling strategies for stable optimization, and the hybrid reward mechanisms to balance evaluation precision and recall.

\subsection{Preliminaries}

The standard RL objective for language models optimizes a KL-regularized expected reward~\citep{schulman2017equivalence}:
\begin{equation}
\max_{\pi_\theta} \mathbb{E}_{x \sim \mathcal{D}, y \sim \pi_{\theta}(\cdot \mid x)} 
\left[ r_{\phi}(x, y) \right] 
- \beta \mathbb{D}_{\text{KL}} \left[ \pi_{\theta}(y \mid x) \,||\, \pi_{\text{ref}}(y \mid x) \right]    
\end{equation}
where $r_{\phi}(x,y)$ denotes the reward for output $y$ given input $x$ from the policy model $\pi_{\theta}$, and $\pi_{\text{ref}}$ represents the reference model for $\mathbb{D}_{\text{KL}}$ regularization.

Unlike prior works wherein the input $x$ typically is a short question, requiring the policy model $\pi_{\theta}$ to generate output $y$ based on its parametric knowledge, we extend this formulation to incorporate an additional long-context $c$. Therefore, the policy model $\pi_{\theta}$ needs to first ground relevant information from $c$, and then provide reasoning chains to solve the question $x$:
\begin{equation}
\max_{\pi_\theta} \mathbb{E}_{\textcolor{blue}{x}, \textcolor{red}{c} \sim \mathcal{D}, y \sim \pi_{\theta}(\cdot \mid \textcolor{blue}{x}, \textcolor{red}{c})} 
\left[ r_{\phi}(\textcolor{blue}{x}, \textcolor{red}{c}, y) \right] 
- \beta \mathbb{D}_{\text{KL}} \left[ \pi_{\theta}(y \mid \textcolor{blue}{x}, \textcolor{red}{c}) \,||\, \pi_{\text{ref}}(y \mid \textcolor{blue}{x}, \textcolor{red}{c}) \right]    
\end{equation}
The context $c$ and the question $x$ are concatenated by the following prompt template:

\begin{table}[!ht]
    \centering
    \caption{Prompt template to concatenate the input \textcolor{red}{context} and \textcolor{blue}{question}.}
    \begin{tabular}{p{13cm}}
        \toprule
        Please read the following text and answer the question below. \\
        <text> \textcolor{red}{context} </text> \textcolor{blue}{question} \\
        Format your response as follows: "Therefore, the answer is (insert answer here)". \\
        \bottomrule
    \end{tabular}
    \label{tab:prompt_template}
\end{table}

\subsection{RL Algorithms}

\textbf{Proxy Policy Optimization (PPO)} To optimize the policy model $\pi_{\theta}$ using the above objective, the PPO~\citep{schulman2017proximal} algorithm proposes to maximize the clipped surrogate objective:
\begin{equation}
\begin{aligned}
\mathcal{J}_\text{PPO}(\theta) & = \mathbb{E}_{x,c \sim \mathcal{D}, y \sim \pi_{\theta_\text{old}}(\cdot\mid x,c)} \\&
\Bigg[\frac{1}{|y|}\sum_{t=1}^{|y|} \min \Bigg( \frac{\pi_{\theta}(y_t\mid x,c,y_{<t})}{\pi_{\theta_{\text{old}}}(y_t\mid x,c,y_{<t})} A_t, \text{clip} \Bigg( \frac{\pi_{\theta}(y_t\mid x,c,y_{<t})}{\pi_{\theta_{\text{old}}}(y_t\mid x,c,y_{<t})}, 1 - \varepsilon, 1 + \varepsilon \Bigg) A_t \Bigg) \Bigg]
\end{aligned}
\end{equation}
where $\pi_{\theta}$ is the current policy model, $\pi_{\theta_\text{old}}$ is the old policy model before updating, $\varepsilon$ is the clipping hyperparameter, and $A_{t}$ is the advantage estimator of the $t$-th token. For a given input context $c$ and question $x$, $\pi_{\theta_{\text{old}}}$ first generate sequential output $y$, and then $A_{t}$ is calculated to update $\pi_{\theta}$ using the generalized advantage estimation (GAE)~\citep{schulman2015high} with a learned value function $V_{\phi}$.

For long-context inputs, the quadratic complexity of attention mechanisms renders value network training computationally prohibitive. Therefore, we opt for the group-relative RL algorithms, including GRPO~\citep{shao2024deepseekmath} and DAPO~\citep{yu2025dapo}, to estimate the advantage through group-normalized rewards instead of an additional value network.

\textbf{Group Relative Policy Optimization (GRPO)} In GRPO, given the context $c$ and the question $x$, the old policy model $\pi_{\theta_{\text{old}}}$ generates a group of $G$ outputs $\{y_{i}\}_{i=1}^{G}$, with rewards $\{r_{i}\}_{i=1}^{G}$ calculated by the reward function. The optimization objective for the policy model $\pi_{\theta}$ becomes:
\begin{equation}
\begin{aligned}
\mathcal{J}_\text{GRPO}(\theta) & = \mathbb{E}_{x, c \sim \mathcal{D}, \{ y_i \}_{i=1}^{G} \sim \pi_{\theta_\text{old}}( \cdot| x,c)} \Bigg[ \frac{1}{G}\sum_{i=1}^{G} \frac{1}{|y_i|}\sum_{t=1}^{|y_i|} \Bigg( \min \Big(\frac{\pi_{\theta}(y_{i,t} | x,c, y_{i,<t})}{\pi_{\theta_\text{old}}(y_{i,t} | x,c, y_{i,<t})} A_{i,t}, \\&
\text{clip} \Big( \frac{\pi_{\theta}(y_{i,t} | x,c, y_{i,<t})}{\pi_{\theta_\text{old}}(y_{i,t} | x,c, y_{i,<t})}, 1 - \varepsilon, 1 + \varepsilon \Big) A_{i,t} \Big) - \beta \mathbb{D}_{\text{KL}}(\pi_{\theta} || \pi_{\text{ref}}) \Bigg) \Bigg]
\end{aligned}
\end{equation}
where the advantage for token $y_{i,t}$ is estimated by normalizing the group-level rewards: \begin{equation}
A_{i,t} = \frac{r_i - \text{mean}(\{r_i\}_{i=1}^G)}{\text{std}(\{r_i\}_{i=1}^G)}
\end{equation} 
In our implementation, we remove the KL term in the GRPO objective to encourage the exploration capacity of the policy model, following the common suggestions in recent works~\citep{OpenReasonerZero2025,Liu2025UnderstandingRT,yu2025dapo}.

\textbf{Decoupled Clip and Dynamic Sampling Policy Optimization (DAPO)} DAPO integrates several techniques to ensure a more stable and efficient RL process: (1) a higher clip threshold to avoid entropy collapse, (2) a dynamic sampling strategy to remove examples with zero reward variance, (3) a token-level loss to mitigate the length bias, and (4) an overlong reward shaping to avoid excessively long outputs. The optimization objective for the policy model $\pi_{\theta}$ is:
\begin{equation}
\begin{aligned}
\mathcal{J}_{\text{DAPO}}(\theta) & = \mathbb{E}_{(x,c)\sim \mathcal{D}, \{y_i\}_{i=1}^G\sim \pi_{\theta_\text{old}}(\cdot\mid x,c)} \Bigg[\frac{1}{\sum_{i=1}^{G}|y_i|}\sum_{i=1}^{G}\sum_{t=1}^{|y_i|} 
\min \Big( \frac{\pi_{\theta}(y_t\mid x,c,y_{<t})}{\pi_{\theta_{\text{old}}}(y_t\mid x,c,y_{<t})} A_{i,t}, \\& 
\text{clip} \Big( \frac{\pi_{\theta}(y_t\mid x,c,y_{<t})}{\pi_{\theta_{\text{old}}}(y_t\mid x,c,y_{<t})}, 1 - {\varepsilon_{\text{low}}}, 1 + {\varepsilon_{\text{high}}} \Big) A_{i,t} \Big) \Bigg]
\\
\end{aligned}
\end{equation}
where $\varepsilon_\text{low}$ and $\varepsilon_\text{high}$ are the low and high clip thresholds, respectively. The dynamic sampling strategy ensures that the examples have a non-zero reward variance, and the token-level loss sets an equal weight for loss calculation for each token in the outputs. The overlong reward shaping set a soft length-award punishment as follows:
\begin{equation}
r_{\text{final}_i} =
\begin{cases}
r_{i}, & |y_{i}| \le L_{\text{max}} - L_{\text{cache}} \\
r_{i} + \frac{(L_{\text{max}} - L_{\text{cache}}) - |y_{i}|}{L_{\text{cache}}}, & L_{\text{max}} - L_{\text{cache}}<|y_{i}|\le L_{\text{max}} \\
r_{i} - 1, & L_{\text{max}} < |y_{i}|
\end{cases}
\label{eq:soft_punish}
\end{equation}
where $L_{\text{max}}$ is the maximum sequence length and $L_{\text{cache}}$ is a buffer zone for gradual length penalization. Different from GRPO, the final advantage for $y_{i,t}$ in DAPO becomes:
\begin{equation}
A_{i,t} = \frac{r_{\text{final}_i} - \text{mean}(\{r_{\text{final}_i}\}_{i=1}^G)}{\text{std}(\{r_{\text{final}_i}\}_{i=1}^G)}
\end{equation}
\subsection{Progressive Context Scaling}
\label{sec:progressive_context_scaling}

Training LRMs for long-context reasoning presents unstable optimization dynamics. To address these issues, we propose a progressive context scaling framework, including a curriculum-guided RL strategy to stabilize short-to-long context optimization, a difficulty-aware retrospective sampling mechanism to prioritize exploration of complex instances, and a warm-up supervised fine-tuning (SFT) phase to provide robust initialization before RL training.

\textbf{Curriculum-Guided Phased Reinforcement Learning} The RL process is divided into $K$ discrete phases, with the target context lengths of $L_k$ for each phase. Starting from an initial input length $L_1$, each subsequent phase increases the input length until reaching the maximum target length $L_K$. During phase $k$, the policy model $\pi_{\theta}$ is trained exclusively on examples that satisfies: 
\begin{equation}
L_{k-1} < |x| + |c| \leq L_k, \quad L_{0} = 0
\end{equation}
where $|x|$ and $|c|$ denote the question length and supporting context length, respectively.

\textbf{Difficulty-Aware Retrospective Sampling} Building on the crucial efficacy of instance difficulty in previous data selection studies~\citep{zhou2023lima,li2024quantity}, we adopt a difficulty-aware retrospective sampling method to strategically incorporate instances from preceding phases. Specifically, we implement importance sampling weighted by difficulty scores to curate retrospective instances:
\begin{equation}
\text{diff}(x,c) = \frac{1}{\text{mean}(\{r_i\}_{i=1}^G)}
\end{equation}
where $\text{diff}(\cdot)$ denotes the difficulty function, quantified as the inverse mean reward $\{r_i\}_{i=1}^G$ from a group of outputs generated by the base model. Lower mean rewards correspond to higher difficulty scores, prioritizing challenging instances during retrospective sampling.

\textbf{Warm-Up Supervised Fine-Tuning} Prior to initiating RL training, we employ a warm-up supervised fine-tuning (SFT) stage to establish a robust initial policy model capable of grounding information from the long-context inputs. This critical preparatory stage ensures the policy model develops fundamental capabilities in context comprehension, reasoning chain generation, and answer extraction before exposure to the instability of RL optimization.  

The SFT process utilizes high-quality demonstrations $\mathcal{D}_{\text{SFT}}$ distilled from a teacher LRM, where each example contains a question $x$, supporting context $c$, and a gold-standard reasoning path $y^*$ with verified correctness. To align with the progressive scaling curriculum, we construct $\mathcal{D}_{\text{SFT}}$ within the initial input length $L_1$ in curriculum-guided RL. The model is trained to minimize the standard negative log-likelihood objective:
\begin{equation}
\mathcal{L}_{\text{SFT}}(\theta) = -\mathbb{E}_{(x,c,y^*) \sim \mathcal{D}_{\text{SFT}}} \frac{1}{|y^*|}\sum_{t=1}^{|y^*|} \log \pi_{\theta}(y_t^* \mid x, c, y_{<t}^*)
\end{equation}
The resulting SFT model serves as the initial policy $\pi_{\theta}$ for RL training, providing stable starting parameters. In Section \ref{sec:ablation}, we demonstrate the effectiveness of the proposed three strategies for stable short-to-long context scaling in reasoning RL.

\subsection{Hybrid Reward Mechanisms}
\label{sec:reward_function_design}

Prior works on short-context reasoning tasks in mathematics, coding, and logical reasoning~\citep{deepscaler2025,el2025competitive,xie2025logic} typically utilize rule-based reward functions that prioritize precision through strict answer matching and format verification to mitigate reward hacking risks~\citep{skalse2022defining}. However, long-context reasoning tasks such as open-domain question answering present unique challenges due to their inherent answer diversity. Overly restrictive rule-based rewards in such contexts risk constraining valid answer variations, potentially compromising overall performance. To address these limitations, we propose a hybrid reward mechanism that combines rule-based verification~\citep{guo2025deepseek} with LLM-as-a-judge~\citep{zheng2023judging}, thereby balancing precision and recall through complementary evaluation.

\textbf{Rule-Based Verification} The rule-based component $\text{r}_{\text{rule}}$ ensures precision by verifying strict adherence to task-specific correctness criteria. For question answering tasks, we first extract the final answer $y_{\text{ans}}$ from model generations $y$ using regular expressions aligned with the structured prompt template in Table~\ref{tab:prompt_template}, and then perform exact string matching against the gold answer $y_{\text{gold}}$:
\begin{equation}
\text{r}_{\text{rule}}(y) = \mathbb{I}(y_{\text{ans}} = y_{\text{gold}})
\end{equation}
where $\mathbb{I}$ represents the indicator function. Notably, we intentionally omit the format reward for answer extraction, as the base model demonstrates sufficient inherent format compliance capabilities; excessive format rewards could oversimplify the learning objective, potentially hindering the model's ability for reasoning chain exploration~\citep{zeng2025simplerlzooinvestigatingtamingzero, jin2025search}.

\textbf{LLM-as-a-Judge} To complement the precision-oriented rule-based component and address potential false negatives in string matching, we introduce an LLM-based evaluator $\text{r}_{\text{LLM}}$ that assesses semantic equivalence between generated and gold answers:
\begin{equation}
\text{r}_{\text{LLM}}(x,y) = \text{LLM}(x,y_{\text{ans}},y_{\text{gold}})
\end{equation}
where the LLM judge produces a binary correctness score based on the evaluation template as follows:

\begin{table}[!ht]
    \centering
    \caption{Prompt template for LLM-as-a-judge to compare the \textcolor{red}{predicted answer} and the \textcolor{green}{gold answer} given the question \textcolor{blue}{quesiton}.}
    \begin{tabular}{p{13cm}}
        \toprule
        You are an expert in verifying if two answers are the same. Your input is a problem and two answers, Answer 1 and Answer 2. You need to check if they are equivalent. Your task is to determine if two answers are equivalent, without attempting to solve the original problem. Compare the answers to verify they represent identical values or meaning, even when written in different forms or notations. Your output must follow the following format: \\
        1) Provide an explanation for why the answers are equivalent or not. \\
        2) Then provide your final answer in the form of: [[YES]] or [[NO]] \\
        Problem: \textcolor{blue}{question} Answer 1: \textcolor{red}{predicted answer} Answer 2: \textcolor{green}{gold answer} \\
        \bottomrule
    \end{tabular}
    \label{tab:llm_judge}
\end{table}

\textbf{Combined Reward Formulation} The integrated reward function combines both rule-based verification and LLM-as-a-judge through maximum selection:
\begin{equation}
r_{\phi}(x, y) = \max(\text{r}_{\text{rule}}(y), \text{r}_{\text{LLM}}(x, y))  
\end{equation}
Given the relative simplicity of answer comparison tasks, we employ a small model, e.g., Qwen2.5-1.5B-Instruct~\citep{yang2024qwen2}, with a temperature of zero for deterministic scoring. This configuration enables efficient reward computation during online RL training while maintaining evaluation reliability.

\section{Experimental Setup}

In our experiments, we employ document question answering (DocQA) as our primary evaluation task for long-context reasoning capabilities, as it inherently requires both contextual grounding and multi-step reasoning. This section details our experimental setup for training and evaluation.

\subsection{Training Datasets}
\textbf{RL Dataset} To construct a challenging RL dataset for verifiable long-context reasoning, we develop {\dataname}, which comprises 1.6K DocQA problems across three reasoning domains: (1) \textbf{Mathematical Reasoning}: We use 600 problems from the DocMath~\citep{zhao2024docmath} dataset, requiring numerical reasoning across long and specialized documents such as financial reports\footnote{For DocMath, we sample 75\% items from each subset from its valid split for training and 25\% for evaluation.}; (2) \textbf{Logical Reasoning}: We employ DeepSeek-R1~\citep{guo2025deepseek} to synthesize 600 multi-choice questions requiring logic analysis of real-world documents spanning legal, financial, insurance, and production domains from our curated collection; (3) \textbf{Multi-Hop Reasoning}: We sample 200 examples from MultiHopRAG~\citep{tangmultihop} and 200 examples from Musique~\citep{trivedi2022musique}, emphasizing cross-document reasoning.

\textbf{SFT Dataset} To establish a robust starting point for RL optimization, we distill 5.3K high-quality question-document-answer triplets through DeepSeek-R1~\citep{guo2025deepseek}. Aligned with recent data curation methods for LRMs~\citep{muennighoff2025s1,ye2025limo}, we clean and filter questions based on quality, complexity, and diversity. Additionally, we control the quality and length of the documents to ensure precise contextual information. In Table \ref{tab:statistics}, we provide the statistics of our RL and SFT datasets.

\vspace{-0.1cm}
\subsection{Training Details}

\textbf{Base Model} In our experiments, we initialize our base model with R1-Distill-Qwen-14B and R1-Distill-Qwen-32B~\citep{guo2025deepseek}, subsequently implementing SFT and RL optimization phases\footnote{We exclude 7B/1.5B variants due to their mathematical reasoning feature inherent from Qwen2.5-Math~\citep{yang2024qwen2}.}.

\textbf{RL Training} As depicted in Section \ref{sec:progressive_context_scaling}, we propose a progressive context scaling mechanism for long-context reasoning RL optimization. Specifically, the training process follows a two-phase curriculum context scaling, with 20K input length $L_1$ in phase I, and 60K input length $L_2$ in phase II. We incorporate difficulty-aware retrospective sampling to maintain the most difficult samples with an average accuracy of zero from phase I to II. The training is conducted on 32xA100-80G GPUs, with a train batch size of 128, a mini batch size of 32, a rollout number of 8, and a learning rate of 2e-6. We set a temperature to 0.7 and a top-p to 0.95 with a maximum output length of 10K for sampling.

\textbf{SFT Training} The input length in the SFT stage is set to 20K. The training is conducted on 32xA100-80G GPUs for 3 epochs, with a train batch size of 128, and a learning rate of 5e-6.

\begin{table}[!t]
\centering
    \caption{Detailed statistics of our train and test datasets. Length is calculated by the Qwen tokenizer.}\label{tab:statistics}
        \resizebox{0.999\textwidth}{!}{
        \begin{tabular}{l|cc|ccccccc}
        \toprule
        \multirow{2}{*}{\textbf{Statistics}} & \multicolumn{2}{c|}{\textbf{Train Dataset}} & \multicolumn{7}{c}{\textbf{Test Dataset}} \\
        ~ & \textbf{SFT} & \textbf{RL} & \textbf{DocMath} & \textbf{Frames} & \textbf{2Wiki} & \textbf{HQA} & \textbf{Musi} & \textbf{NarQA} & \textbf{Qasp} \\
        \midrule
        \# Examples & 5,305 & 1,591 & 200 & 824 & 200 & 200 & 200 & 200 & 200 \\
        Avg. Length & 13,064 & 11,437 & 17,645 & 15,756 & 7,530 & 13,431 & 16,327 & 29,887 & 5,074 \\
        Max. Length & 20,003 & 59,559 & 176,285 & 117,131 & 17,035 & 17,640 & 17,883 & 65,357 & 21,927 \\
        \bottomrule
    \end{tabular}
    }
    \vspace{-0.3cm}
\end{table}

\vspace{-0.1cm}
\subsection{Evaluation Details}

\textbf{Benchmarks} We conduct evaluation on seven long-context DocQA benchmarks, including multi-hop reasoning benchmarks\footnote{We use the data from LongBench~\citep{bai-etal-2024-longbench} for 2WikimultihopQA, HotpotQA, Musique, NarrativeQA, and Qasper.} such as 2WikiMultihopQA~\citep{ho2020constructing}, HotpotQA~\citep{yang2018hotpotqa}, Musique~\citep{trivedi2022musique}, NarrativeQA~\citep{kovcisky2018narrativeqa}, Qasper~\citep{dasigi2021dataset}, and Frames~\citep{krishna2024fact} as well as mathematical reasoning benchmarks like DocMath~\citep{zhao2024docmath}. We report the maximum of exact match and LLM-judged accuracy as the final score, aligned with the reward function in Section \ref{sec:reward_function_design}. We use DeepSeek-V3~\citep{liu2024deepseek} as the judge model with a temperature of 0.0 to provide a reliable evaluation. The benchmark statistics are shown in Table \ref{tab:statistics}.

\textbf{Configurations} We evaluate our long-context LRMs with a maximum input length of 120K and output length of 10K. For the proprietary LRMs with a limited context length, we set the maximum input length to 50K. We conduct a zero-shot evaluation with a temperature of 0.7 and a top-p of 0.95.

\vspace{-0.1cm}
\subsection{Baselines}  
We compare {\modelname} against the following state-of-the-art LRMs.

\textbf{Proprietary LRMs} OpenAI-o1-preview~\citep{openai2024openaio1card}, Claude-3.7-Sonnet-Thinking~\citep{claude-3.7}, OpenAI-o3-mini~\citep{openai2025openaio3card}, Qwen3-Plus~\citep{qwen3}, QwQ-Plus~\citep{qwq-32b}, and Gemini-2.0-Flash-Thinking~\citep{gemini-thinking}.

\textbf{Open-Source LRMs} DeepSeek-R1~\citep{guo2025deepseek}, Qwen3-235B-A22B~\citep{qwen3}, Qwen3-32B~\citep{qwen3}, QwQ-32B~\citep{qwq-32b}, R1-Distill-Qwen-32B~\citep{guo2025deepseek}, and R1-Distill-Qwen-14B~\citep{guo2025deepseek}.

\section{Experimental Results}

\subsection{Main Results}

\begin{table}[!t]
\centering
    \caption{Main results across seven long-context DocQA benchmarks. We highlight the \textbf{\textcolor{red}{top-1}} and \textbf{top-3} performance. $\Delta$ indicates the performance \textcolor{blue}{gains} and \textcolor{green}{declines} compared to the base models.}\label{tab:main}
        \resizebox{0.999\textwidth}{!}{
        \begin{tabular}{lcccccccc}
        \toprule
        \textbf{Models} & \textbf{DocMath} & \textbf{Frames} & \textbf{2Wiki} & \textbf{HQA} & \textbf{Musi} & \textbf{NarQA} & \textbf{Qasp} & \textbf{Avg.} \\
        \midrule
        \multicolumn{9}{c}{\textbf{Proprietary LRMs}} \\ \midrule
        OpenAI-o1-preview & 64.5 & \textcolor{red}{\textbf{80.8}} & 87.5 & 83.5 & \textbf{69.0} & \textcolor{red}{\textbf{68.0}} & \textbf{57.0} & \textcolor{red}{\textbf{72.9}} \\
        Claude-3.7-Sonnet-Thinking & \textbf{67.5} & 70.9 & 86.5 & \textbf{84.4} & 68.3 & \textbf{61.5} & 56.0 & \textbf{70.7} \\
        OpenAI-o3-mini & 66.5 & \textbf{75.5} & 86.5 & 83.5 & 66.5 & 59.0 & 55.0 & 70.4 \\
        Qwen3-Plus & 66.0 & 73.6 & \textbf{90.5} & 82.4 & \textbf{69.8} & 57.5 & 52.5 & 70.3 \\
        QwQ-Plus & 64.5 & 73.5 & 89.0 & 81.0 & 66.5 & \textbf{62.0} & 53.5 & 70.0 \\
        Gemini-2.0-Flash-Thinking & 63.0 & 69.8 & 82.9 & 79.5 & 62.5 & 57.0 & 45.5 & 65.7 \\ \midrule
        \multicolumn{9}{c}{\textbf{Open-Source LRMs}} \\ \midrule
        DeepSeek-R1 & 66.0 & \textbf{79.6} & 89.9 & 82.5 & \textcolor{red}{\textbf{74.5}} & 59.5 & 53.0 & \textbf{72.1} \\
        Qwen3-235B-A22B & \textbf{67.5} & 74.6 & \textcolor{red}{\textbf{91.5}} & \textbf{84.4} & 63.3 & 60.0 & 53.0 & 70.6 \\
        QwQ-32B & 59.5 & 72.9 & \textbf{90.5} & 78.5 & 66.0 & 58.0 & \textbf{57.5} & 69.0 \\
        Qwen3-32B & 58.0 & 70.0 & 87.0 & 83.4 & 62.8 & 57.5 & 56.0 & 67.8 \\
        R1-Distill-Qwen-32B & 62.5 & 67.0 & 84.0 & 80.5 & 61.0 & 54.0 & 50.0 & 65.6 \\
        R1-Distill-Qwen-14B & 61.0 & 64.2 & 87.0 & 77.5 & 58.0 & 51.0 & 51.0 & 64.2 \\ \midrule
        \multicolumn{9}{c}{\textbf{Ours Methods}} \\ \midrule
        \rowcolor{Gray} R1-Distill-Qwen-14B-SFT & 60.0 & 65.7 & 88.5 & 80.5 & 60.0 & 52.0 & 48.5 & 65.0 \\
        \quad \textit{$\Delta$ to R1-Distill-Qwen-14B} & (\textcolor{green}{-1.0}) & (\textcolor{blue}{+1.5}) & (\textcolor{blue}{+1.5}) & (\textcolor{blue}{+3.0}) & (\textcolor{blue}{+2.0}) & (\textcolor{blue}{+1.0}) & (\textcolor{green}{-2.5}) & (\textcolor{blue}{+0.8})\\
        \rowcolor{Gray} {\modelname}-14B-GRPO & 65.0 & 68.7 & 88.5 & \textcolor{red}{\textbf{86.5}} & 63.5 & 53.5 & 51.5 & 68.2 \\
        \quad \textit{$\Delta$ to R1-Distill-Qwen-14B} & (\textcolor{blue}{+4.0}) & (\textcolor{blue}{+4.5}) & (\textcolor{blue}{+1.5}) & (\textcolor{blue}{+9.0}) & (\textcolor{blue}{+5.5}) & (\textcolor{blue}{+2.5}) & (\textcolor{blue}{+0.5}) & (\textcolor{blue}{+4.0})\\
        \rowcolor{Gray} {\modelname}-14B-DAPO & 65.5 & 67.4 & 89.0 & \textbf{84.0} & 63.0 & 57.0 & 52.5 & 68.3 \\
        \quad \textit{$\Delta$ to R1-Distill-Qwen-14B} & (\textcolor{blue}{+4.5}) & (\textcolor{blue}{+3.2}) & (\textcolor{blue}{+2.0}) & (\textcolor{blue}{+6.5}) & (\textcolor{blue}{+5.0}) & (\textcolor{blue}{+6.0}) & (\textcolor{blue}{+1.5}) & (\textcolor{blue}{+4.1})\\ \midrule
        \rowcolor{Gray} R1-Distill-Qwen-32B-SFT & 65.0 & 71.6 & 87.0 & 80.5 & 65.5 & 57.5 & 54.0 & 68.7 \\
        \quad \textit{$\Delta$ to R1-Distill-Qwen-32B} & (\textcolor{blue}{+2.5}) & (\textcolor{blue}{+4.6}) & (\textcolor{blue}{+3.0}) & (\textcolor{blue}{+0.0}) & (\textcolor{blue}{+4.5}) & (\textcolor{blue}{+3.5}) & (\textcolor{blue}{+4.0}) & (\textcolor{blue}{+3.2})\\
        \rowcolor{Gray} {\modelname}-32B-GRPO & \textcolor{red}{\textbf{68.0}} & 72.2 & 87.0 & 82.0 & 66.0 & 61.0 & 56.0 & 70.3 \\
        \quad \textit{$\Delta$ to R1-Distill-Qwen-32B} & (\textcolor{blue}{+5.5}) & (\textcolor{blue}{+5.2}) & (\textcolor{blue}{+3.0}) & (\textcolor{blue}{+1.5}) & (\textcolor{blue}{+5.0}) & (\textcolor{blue}{+7.0}) & (\textcolor{blue}{+6.0}) & (\textcolor{blue}{+4.7})\\
        \rowcolor{Gray} {\modelname}-32B-DAPO & \textbf{67.5} & 70.1 & \textbf{90.5} & 83.0 & \textbf{69.0} & 56.0 & \textcolor{red}{\textbf{58.5}} & \textbf{70.7} \\
        \quad \textit{$\Delta$ to R1-Distill-Qwen-32B} & (\textcolor{blue}{+5.0}) & (\textcolor{blue}{+3.1}) & (\textcolor{blue}{+6.5}) & (\textcolor{blue}{+2.5}) & (\textcolor{blue}{+8.0}) & (\textcolor{blue}{+2.0}) & (\textcolor{blue}{+8.5}) & (\textcolor{blue}{+5.1})\\
        \bottomrule
    \end{tabular}
    }
\end{table}

\begin{figure}[!t]
    \centering
    \includegraphics[width=0.999\textwidth]{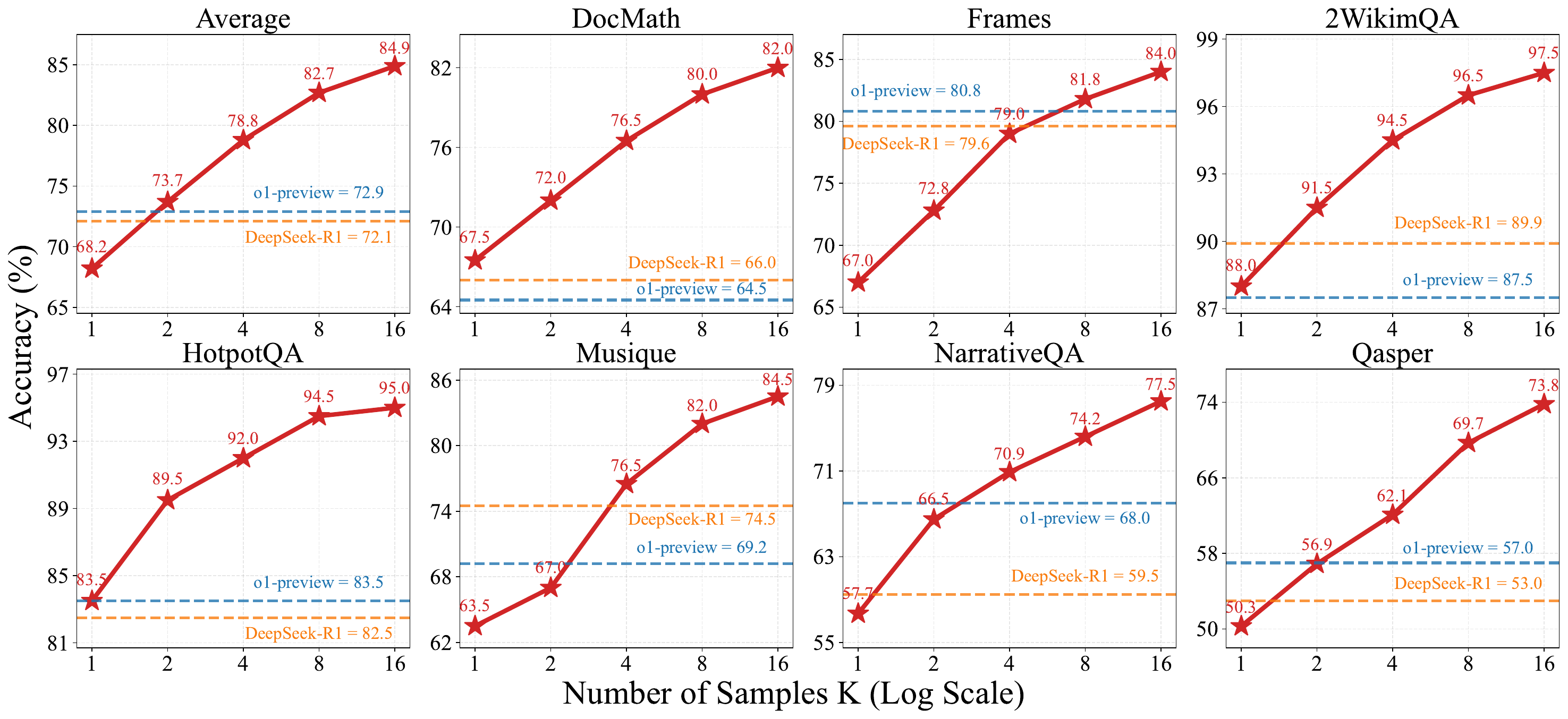}
    \caption{Pass@K rates of {\modelname}-14B with different sample numbers across all benchmarks. We show that {\modelname}-14B surpasses DeepSeek-R1 with a small sampling number.}
    \label{fig:test_time_scaling}
    \vspace{-0.3cm}
\end{figure}

Table \ref{tab:main} presents the overall performance of {\modelname} across seven long-context document question answering (DocQA) benchmarks. The key findings are as follows:

\textbf{Limited Efficacy of SFT for Long-Context Reasoning}. Since the base model, R1-Distill-Qwen, is primarily optimized for short-context reasoning tasks in mathematics, coding, and scientific domains, we conduct SFT to adapt it for long-context reasoning before RL training, as outlined in Section \ref{sec:progressive_context_scaling}. Despite this intervention, the SFT model only shows an average gain of 0.8 points on 14B and 3.2 points on 32B. Furthermore, the improvements exhibit significant variability across benchmarks, suggesting limited generalizability of the SFT approach to long-context reasoning scenarios.

\textbf{Significant Improvements via RL Integration}. Through the integration of RL, {\modelname} exhibits remarkable advancements in long-context reasoning performance. Notably, {\modelname}-14B achieves an average improvement of 4.1 and 4.0 points over the base model with DAPO and GRPO, surpassing the 0.4 points improvement of the SFT baseline by a significant margin. Furthermore, when scaling to 32B base model, {\modelname}-32B even demonstrates a 5.1 and 4.7 points performance increase with DAPO and GRPO. These results highlight the advanced capacity of RL approaches in refining the output distribution to address intricate, context-dependent reasoning problems through group-relative advantage estimation and incentives for on-policy sampled outputs.

\textbf{Leading Performance among Flagship LRMs}. Our evaluation demonstrates that {\modelname} achieves superior performance compared to leading proprietary and open-source LRMs. Specifically, {\modelname}-14B achieves an average score of 68.3, surpassing Gemini-2.0-Flash-Thinking, R1-Distill-Qwen-32B, and Qwen3-32B, while mathcing the performance of QwQ-32B. Moreover, {\modelname}-32B achieves an average score of 70.7, exceeding the performance of QwQ-Plus, Qwen3-Plus, Qwen3-235B-A22B, and OpenAI-o3-mini, even comparable to Claude-3.7-Sonnet-Thinking, demonstrating leading performance among state-of-the-art flagship LRMs. 

\textbf{Additional Enhancements by Test-Time Scaling}. We further conduct experiments to analyze the test-time scaling performance of {\modelname}. Following established works~\citep{guo2025deepseek,wen2025light}, we generate 16 candidate outputs per input question and evaluate Pass@K to quantify exploratory capability across all benchmarks. As illustrated in Figure \ref{fig:test_time_scaling}, {\modelname}-14B exhibits consistent performance enhancements with increased sampling scales. Notably, {\modelname}-14B demonstrates remarkable gains, even surpassing DeepSeek-R1 and OpenAI-o1-preview with a small sample size. Specifically, it achieves an average Pass@2 rate of 73.7 across all benchmarks, outperforming both DeepSeek-R1's 72.1 and OpenAI-o1-previews's 72.9, highlighting the efficacy of test-time scaling. Moreover, the significant gap between Pass@K and Pass@1 metrics indicates further potential for RL training to better bridge the transition from diverse exploration to optimal exploitation.

\begin{figure}[!t]
    \centering
    \includegraphics[width=0.999\textwidth]{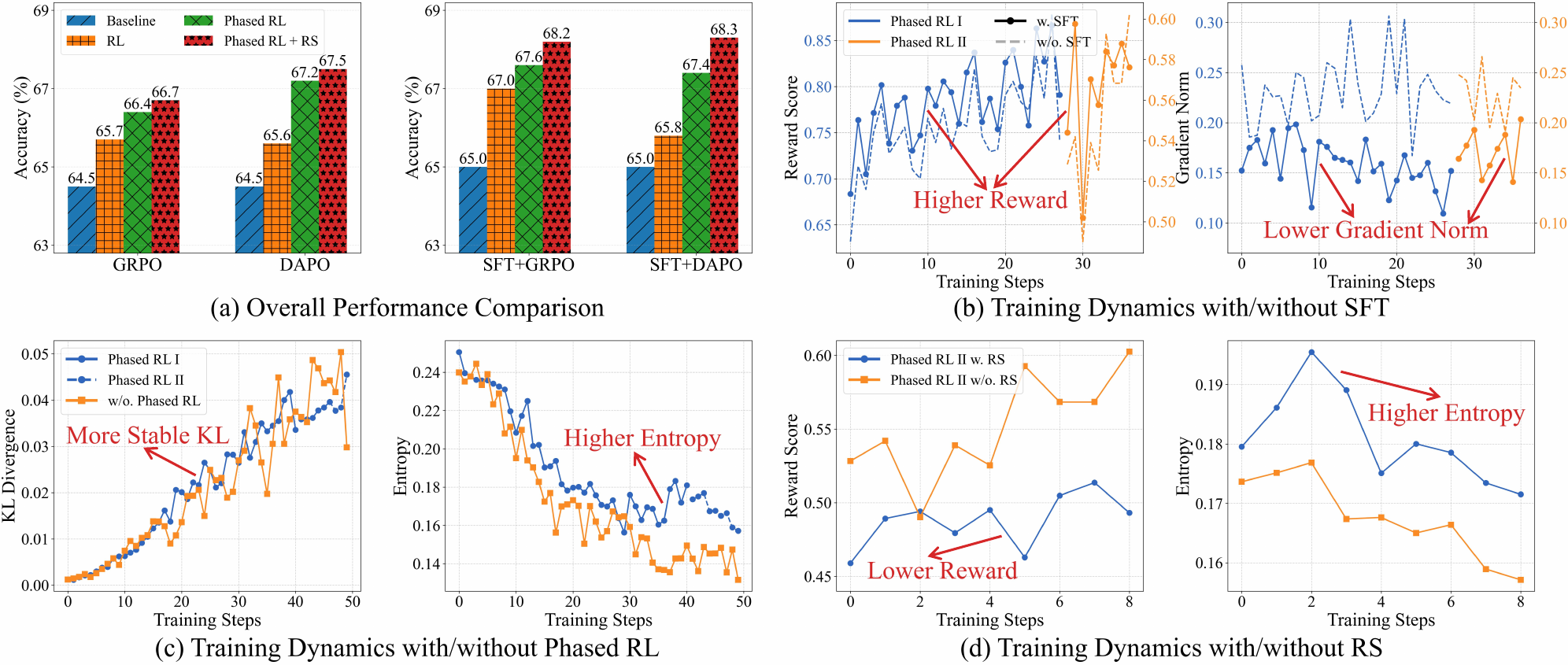}
    \caption{Ablation studies of progressive context scaling strategy, where ``Baseline'' refers to the base or SFT model before RL training, ``RL'' refers to the naive single-stage RL, and  ``Phased RL'' refers to the curriculum-guided phased RL. ``RS'' refers to the difficulty-aware retrospective sampling.}
    \label{fig:ablation_studies}
    \vspace{-0.3cm}
\end{figure}

\subsection{Ablation Studies}
\label{sec:ablation}

In this section, we conduct ablation studies to investigate the key components in {\modelname} that enable successful progressive context scaling for long-context reasoning RL, including warm-up supervised fine-tuning, curriculum-guided phased reinforcement learning, and difficulty-aware retrospective sampling, with the experimental results shown in Figure \ref{fig:ablation_studies}.

\textbf{Warm-up Supervised Fine-tuning}. To illustrate the influence of warm-up SFT, we first evaluate the overall performance of models trained with and without this preparatory phase across seven benchmarks, using various RL algorithms and training strategies. As illustrated in Figure \ref{fig:ablation_studies} (a), integration warm-up SFT yields significant performance improvements in all experimental setups. To further explore the mechanism of warm-up SFT in RL dynamics, Figure \ref{fig:ablation_studies} (b) tracks the reward scores and gradient norm during training. The results reveal that warm-up SFT not only accelerates reward improvements but also sustains lower gradient norm during different RL phases, validating its capacity to prioritize performance gains over format alignment when transitioning models from short-context to long-context reasoning tasks. These findings highlight the necessity of integrating SFT as a precursor to providing a robust and efficient initialization for RL training.

\textbf{Curriculum-Guided Phased Reinforcement Learning}. As shown in Figure \ref{fig:ablation_studies} (a), we conduct a comparative analysis between naive single-stage RL and the proposed curriculum-guided phased RL, with different training configurations: GRPO, DAPO, SFT + GRPO, and SFT + DAPO. The results demonstrate that our phased RL methodology achieves substantial performance improvements. We also note that this improvement is less pronounced when models are initialized with SFT,  suggesting that warm-up training partially compensates for curriculum design. Further analysis in Figure \ref{fig:ablation_studies} (c) reveals that single-stage RL exhibits heightened instability, as demonstrated by fluctuating KL divergence and entropy collapse. These results confirm the pivotal role of curriculum-guided phased training in the stable policy evolution from short-context to long-context reasoning RL.

\textbf{Difficulty-Aware Retrospective Sampling}. To maintain a wild exploration of hard examples, we introduce a difficulty-aware retrospective sampling strategy to integrate a subset of hard samples from prior training phases into the current training data. As illustrated in \ref{fig:ablation_studies} (a), this strategy yields further performance enhancements with phased RL. Notably, in Figure \ref{fig:ablation_studies} (d), despite undergoing phase I RL training, these retained hard examples also lead to significantly lower reward and higher policy entropy, which incentivize the policy model to augment the exploration process.

\begin{wrapfigure}[]{r}{0.40\textwidth}
\vspace{-0.7cm}
\begin{center}
\includegraphics[width=0.40\textwidth]{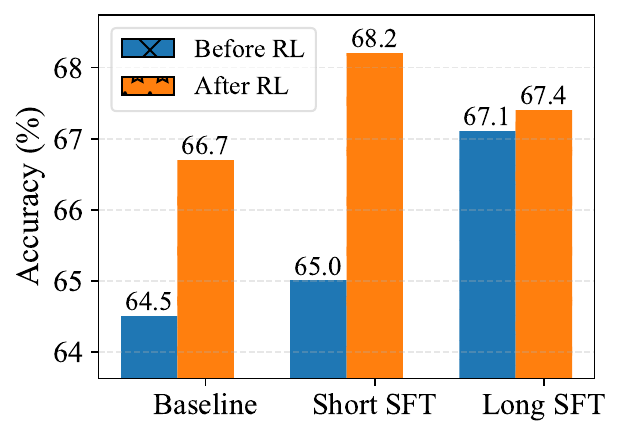}
\end{center}
\vspace{-0.4cm}
\caption{Comparison between different models before and after RL, where ``Baseline'' denotes the base model, ``Short SFT'' denotes the short-context SFT model, and ``Long SFT'' denotes the long-context SFT model.}
\vspace{-0.4cm}
\label{fig:sft_and_rl}
\end{wrapfigure}

\vspace{-0.1cm}
\subsection{Additional Analysis}
\label{sec:additional_analysis}

In this section, we investigate the questions pertaining to the development of long-context LRMs, focusing on the trade-off between SFT and RL in optimizing long-context reasoning capabilities, and the emergence and dynamics of long-context reasoning behaviors during training.

\textbf{Trade-off between SFT and RL in Optimization}. As discussed in Section \ref{sec:ablation}, SFT offers a robust initialization for RL training. However, given that the initial SFT phase in our experiments relied on short-context training data, a critical question arises regarding the role of long-context SFT and its impact on RL. To this end, we train a long-context SFT model using 10K context-question-answer triplets distilled from DeepSeek-R1, maintaining the same data distribution as the short-context SFT phase. This long-context SFT model serves as the starting point for single-stage RL training, without progressive context scaling due to its inherent long-context capability.

As shown in Figure \ref{fig:sft_and_rl}, the long-context SFT model surpasses both the base model by 2.6 points and the short-context SFT model by 2.1 points. Despite the requirement for more training data, SFT offers distinct practical advantages, including reduced computational complexity, minimal infrastructure demands, and diminished reliance on specialized technical expertise, thereby positioning it as an economical strategy for performance enhancement~\citep{guo2025deepseek}. However, further RL applied to the long-context SFT model yields marginal improvements, with only 0.3 points gains and a 67.4 final score, significantly underperforming the 3.2 points improvements and a 68.2 final score achieved when RL is applied to the short-context SFT model. These results highlight two insights for long-context LRM development: (1) SFT and RL exhibit distinct yet complementary purposes—SFT achieves acceptable performance with less effort, whereas RL is indispensable for attaining optimal results; (2) Maximizing performance necessitates prioritizing RL over SFT, as excessive focus on SFT risks trapping models in local optima, thereby constraining opportunities for RL improvements.

\textbf{Emergence and Dynamics of Long-Context Reasoning Behaviors}. The reasoning behaviors critically shape LRMs' reasoning trajectories and rewards~\citep{guo2025deepseek,OpenReasonerZero2025}. To investigate these dynamics, we follow recent studies~\citep{zeng2025simplerlzooinvestigatingtamingzero,Liu2025UnderstandingRT} to analyze the evolution of reasoning behaviors during SFT and RL training. Specifically, we use DeepSeek-V3~\citep{liu2024deepseek} to extract and track the shifts of the average count of four core reasoning behaviors over training steps, including long-context specific \textit{grounding} and three general reasoning behaviors: \textit{subgoal setting}, \textit{backtracking}, and \textit{verification}:

\begin{itemize}[leftmargin=1em]
    \item \textbf{Grounding}: The model recalls related information in the long context to support subsequent reasoning, \textit{e.g., ``Let me look through the provided text to find...''}.
    \item \textbf{Subgoal Setting}: The model decomposes complex questions into multiple manageable subgoals to solve them step-by-step, \textit{e.g., ``To solve this, we first need to...''}.
    \item \textbf{Backtracking}: The model identifies errors in generations and go back to revise its approach iteratively, \textit{e.g., ``This approach won't work because...''}.
    \item \textbf{Verification}: The model validates the predicted answers systematically to ensure solution correctness with self-reflection, \textit{e.g., ``Let's verify this result by...''}.
\end{itemize}

\begin{figure}[!t]
    \centering
    \includegraphics[width=0.999\textwidth]{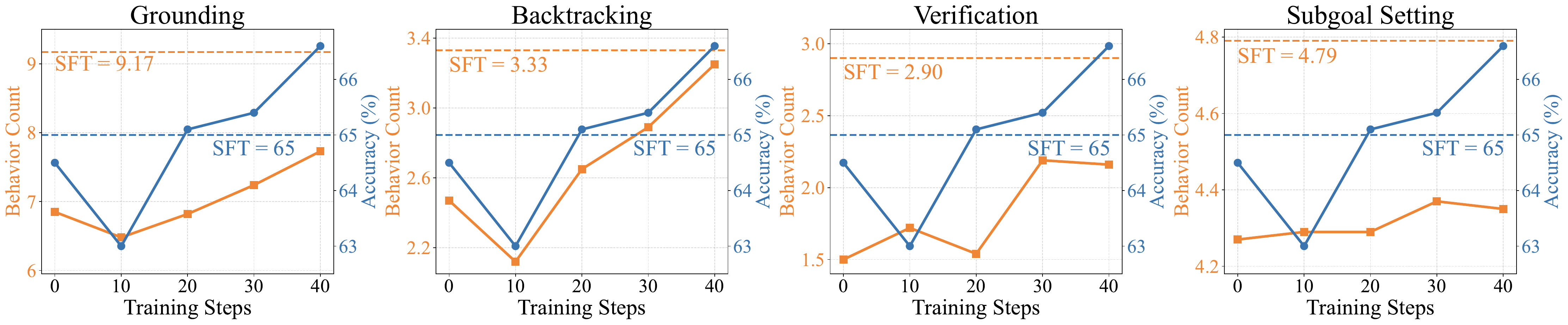}
    \caption{The change in reasoning behavior over training steps. We focus on four core reasoning behaviors, including long-context specific \textit{grounding} and three general reasoning strategies: \textit{subgoal setting}, \textit{backtracking}, and \textit{verification}. During RL training, we show that all the \textcolor{orange}{behaviors} increase progressively with the corresponding \textcolor{blue}{performance} gains. However, despite SFT leading to significantly increased reasoning behaviors, these efforts fail to improve the final performance.}
    \vspace{-0.3cm}
    \label{fig:behaviors}
\end{figure}

The results in Figure \ref{fig:behaviors} reveal three insights: (1) All LRMs exhibit marked reasoning behaviors, with long-context grounding occurring most frequently, underscoring its effectiveness in managing contextual dependencies during reasoning. (2) RL training amplifies these behaviors progressively, correlating with significant performance gains, suggesting RL's efficacy in refining the output space to prioritize reasoning patterns conducive to accurate solutions. (3) In contrast, while SFT models demonstrate increased reasoning behaviors, these adjustments fail to transform into performance improvements, likely due to SFT’s inherent reliance on imitation learning, which prioritizes superficial pattern alignment over substantive reasoning skill development~\citep{zhou2023lima,linunlocking}.

\section{Conclusion and Future Work}

In this study, we explore the development of long-context LRMs with robust contextual grounding and reasoning capabilities through reinforcement learning (RL). We first propose the paradigm of \textit{long-context reasoning RL} and identify \textit{suboptimal training efficiency} and \textit{unstable optimization process}. To address these challenges, we present {\modelname}, a progressive context scaling RL framework designed to bridge the gap between short-context proficiency and long-context generalization. Specifically, the training process begins with a warm-up SFT, followed by a curriculum-guided phased RL, with a difficulty-aware retrospective sampling strategy. Experiments across seven long-context document question-answering benchmarks demonstrate that {\modelname} achieves leading performance among state-of-the-art proprietary LRMs. Specifically, {\modelname}-14B outperforms Gemini-2.0-Flash-Thinking and Qwen3-32B, while {\modelname}-32B further surpasses OpenAI-o3-mini, Qwen3-235B-A22B, and even matches Claude-3.7-Sonnet-Thinking. Our analysis yields three key insights for long-context reasoning RL, including the pivotal role of progressive context scaling in enabling stable adaptation, the necessity of prioritizing RL for optimal performance, and the increase of long-context reasoning behaviors during RL training for performance improvements.

Future work should prioritize three key avenues to advance long-context LRMs. First, scaling real-world tasks, like automated scientific research and long video analysis, will provide appropriate environments to enhance long-context comprehension and decision-making capabilities. Second, developing advanced architectures is essential, including optimized attention mechanisms, e.g., linear and sparse attention, and efficient infrastructures, e.g., asynchronous actor rollout and parameter updating. Third, rethinking long-context RL paradigms, such as transitioning from token-level to turn-level markov decision process (MDP), might enable the breakdown of long-context into sequential interactions and optimizing them iteratively, paving the way for infinite-context RL systems.

\definecolor{case1_color}{RGB}{230, 238, 255}
\definecolor{case2_color}{RGB}{253, 231, 205}

\section{Case Study}  

To demonstrate {\modelname}'s capabilities, we show two illustrative examples from our analysis.  

Case \texttt{complong-testmini-183}: When evaluating {\modelname}-14B against R1-Distill-Qwen-14B and R1-Distill-Qwen-14B-SFT, distinct behavioral differences emerge. In this instance, R1-Distill-Qwen-14B is misled by the detailed \textit{``Interest is payable semi–annually beginning October 15, 2011''}. It grounds \textit{irrelevant temporal} and \textit{financial information}, and consequently \textit{miscalculates the first-year interest payment}. Although R1-Distill-Qwen-14B-SFT is fine-tuned on high-quality SFT triplets distilled from DeepSeek-R1, it still fails to resolve this issue. Instead, it demonstrates persistent limitations by \textit{entering a loop of over-analysis on unrelated documents}. The \textit{uncertainty} and \textit{self-doubting} behavior eventually causes it to \textit{exhaust the maximum generation limit} (10,000 tokens) without providing the final answer. In contrast, while {\modelname}-14B initially exhibits similar distractions, it quickly \textit{engages in effective self-reflection}. Through \textit{timely verification and backtracking}, it successfully \textit{filters out the irrelevant details and arrives at the correct answer}. 

Case \texttt{complong-testmini-265}: This example illustrates how long-context reasoning behaviors (e.g. \textit{grounding}, \textit{backtracking}, \textit{verification}, \textit{subgoal setting}) emerge and interplay in practice, complementing our behavior evolution analysis in Figure~\ref{fig:behaviors}. This task requires the model to \textit{infer the total interest accrued over two years based on dense financial filings}. Here, the model \textit{correctly computes that the total interest is \$980,000} based on a principal of \textit{\$4,900,000 at 10\% annual interest over two years}. This success is attributed to the emergence of multiple reasoning behaviors, showcasing how long-context reasoning RL enhances not only improves answer accuracy but also fosters the development of internal cognitive behaviors during reasoning.

\begin{tcolorbox}[
  title=DocMath complong-testmini-183,
  breakable,   
  fonttitle=\bfseries,
  enhanced,                        
  colback=case1_color!10,           
  colbacktitle=case1_color,         
  coltitle=black,                 
  colframe=case1_color!80!black,    
  coltext=black,                  
  boxrule=0.5pt,
  arc=2mm
]
\label{tab:case1}
\small
\textbf{Ground Truth:} 32.4 

\textbf{Prompt:} Please read the following text and answer the question below.

\colorbox{case1_color}{<text>}

{\centering ...\textit{127Docs}... \par}

128: On March 22, 2011, we issued \$300.0 million in aggregate principal amount of 8.0\% senior unsecured notes due 2019 (the "Notes") at an offering price equal to 100\% of par. The Notes were sold in a private placement to eligible purchasers in a transaction exempt from the registration requirements of the Securities Act of 1933, as amended.

129: We received net proceeds of \$291.6 million, after deducting the discount of \$7.5 million and offering expenses of \$0.9 million. We used the net proceeds to repay indebtedness under our existing credit facility. The discount and the offering expenses are being amortized over the life of the Notes. The amortization is included in "Interest expense" on our unaudited condensed consolidated statements of operations.

130: The Notes were issued under an indenture dated March 22, 2011, (the "Indenture"), mature April 15, 2019, and bear interest at 8.0\%. Interest is payable semi–annually beginning October 15, 2011. The Notes are general unsecured obligations and are effectively junior in right of payment to any of our secured indebtedness to the extent of the value of the collateral securing such indebtedness.
 
{\centering ...\textit{271Docs}... \par}

\colorbox{case1_color}{</text>}

What is the total capital cost, combining both the offering cost of the senior notes and interest payment in the first year, in million dollars?

Format your response as follows: "Therefore, the answer is (insert answer here)".

\tcbline
~
\\*[0.5mm]
\textbf{R1-Distill-Qwen-14B:} 
\\*[3mm]
\textbf{Response (2166 tokens):} 
\\*[3mm]
\colorbox{case1_color}{<think>}

Alright, so I need to figure out the total capital cost for EV Energy Partners, L.P., combining both the offering cost of the senior notes and the interest payment in the first year. Let me go through the provided financial statements step by step to find all the relevant information.

First, I’ll look for the senior notes offering costs. [...] The offering expenses were \$0.9 million, and the discount was \$7.5 million. [...] So the total
offering costs (discount and expenses) are \$7.5 + \$0.9 = \$8.4 million.

Next, I need to find the interest payment for the first year. The senior notes have an 8\% annual interest
rate and are due in 2019. They were issued in March 2011, so the first interest period would be from
March to September 2011, but the interest is actually payable semi-annually starting in October 2011.

[...] 

\textcolor{red}{Since the notes were outstanding for six months, the semi-annual interest would be 8\% / 2 = 4\% of the principal. Therefore, the interest for the first year would be 4\% of \$300 million, which is \$12 million.}

[...] 

The interest is semi-annual, so the first payment would be in October 2011, and the second in April
2012. So for the first year (2011), they would have paid \$12 million in interest on the senior notes. 

[...] 

So combining the offering cost (\$8.4 million) and the interest payment for the first year (\$12 million),
the total capital cost is \$8.4 + \$12 = \$20.4 million.

[...]

\colorbox{case1_color}{</think>}

The total capital cost includes the offering costs and the interest payment for the first year. The offering
costs for the senior notes were \$8.4 million, and the interest expense for the first year (2011) was \$12
million. Therefore, the total capital cost is \$8.4 million + \$12 million = \$20.4 million. 

Therefore, the answer is \boxed{\$20.4} million.
\tcbline
~
\\*[0.5mm]
\textbf{Grounding Behavior:} 
\\*[3mm]
\textbf{Instances of Grounding Behavior:}

\begin{enumerate}[leftmargin=0.5cm]
    \item \textbf{Explicit Grounding}: "Let me go through the provided financial statements step by step to find all the relevant information."
    
    \item \textbf{Explicit Grounding}: "In the Notes section, under NOTE 5: FAIR VALUE MEASUREMENTS, I see that on March 22, 2011, they issued \$300 million in senior notes."
    
    \item \textbf{Explicit Grounding}: "Wait, but the offering cost is mentioned in NOTE 9: OWNER'S EQUITY."
    
    \item \textbf{Explicit Grounding}: "Looking at the balance sheet, under June 30, 2011, the long-term debt is \$480,183 thousand, which is \$480.183 million."
    
    \item \textbf{Explicit Grounding}: "Wait, in NOTE 7: LONG-TERM DEBT, under the credit facility, it says that borrowings bear interest at a floating rate based on LIBOR."
    
    \item \textbf{Grounding Verification}: "Wait, the financial statements show interest expense for the three months ended June 30, 2011 as \$8,124 thousand, which is \$8.124 million."
    
    \item \textbf{Grounding Verification}: "Wait, but in the condensed statements, the interest expense for the three months is \$8.124 million, and for six months, it's \$13,283 thousand."
    
    \item \textbf{Grounding Verification}: "Wait, the notes were issued in March 2011, so from March 2011 to March 2012 is the first year."
    
    \item \textbf{Grounding Verification}: "Wait, in the financial statements, the interest expense for the six months ended June 30, 2011 is \$13.283 million, which includes the senior notes' interest for three months and the credit facility's interest for the entire six months."
    
    \item \textbf{Grounding Verification}: "Wait, the senior notes were issued in March 2011, so the first interest period is from March 22, 2011 to September 22, 2011, which is about six months, and the interest would be 4\% of \$300 million, which is \$12 million."
    
    \item \textbf{Grounding Verification}: "Wait, but the offering cost is a one-time expense, and the interest is an ongoing expense."
    
    \item \textbf{Grounding Verification}: "Wait, but the question is about the total capital cost, which usually includes both the initial costs and the interest expense over the period."
\end{enumerate}

Each of these instances involves either explicit references to the document or verification steps that lead back to the document, thus counting as grounding behavior.
\tcbline
~
\\*[0.5mm]\\*[3mm]
\textbf{R1-Distill-Qwen-14B-SFT:} 
\\*[3mm]
\textbf{Response (10000 tokens):} 
\\*[3mm]
\colorbox{case1_color}{<think>}

Okay, let's tackle this question. [...]

First, I need to find where the offering costs for the senior notes are mentioned. [...] So the total offering costs here are \$7.5 million (discount) plus \$0.9 million (expenses), which adds up to \$8.4 million.

Next, the interest payment for the first year. The senior notes were issued in March 2011 with a 8.0\% annual interest rate. The total principal amount is \$300 million. [...] \textcolor{red}{However, the notes mention that interest is payable semi-annually starting October 15, 2011.} [...] Wait, the notes were issued on March 22, 2011, and the first interest payment is on October 15, 2011. 

[...]

The interest expense for the three months ended June 30, 2011, under "Interest expense" is \$8,124 thousand. 

[...]

In the Condensed Consolidated Statements of Operations for the six months ended June 30, 2011, the interest expense is \$13,283 thousand. 

[...] 

In Note 7, under the senior notes, it says: "The discount and the offering expenses are being amortized over the life of the Notes. The amortization is included in 'Interest expense' on our unaudited condensed consolidated statements of operations." So the total initial costs for the senior notes are \$7.5 million + \$0.9 million = \$8.4 million. These are amortized over the life of the notes, so each year, \$8.4 million / 8 = \$1.05 million per year. For the first year (2011), that would be \$1.05 million.
The question is about the interest payment in the first year. 

[...] 

The senior notes [...] bear 8\% interest, payable semi-annually. The first payment is on October 15, 2011 [...]. So in the first year (2011), there's only one payment in October. 

[...] 

The coupon rate is 8\%, so each payment is 4\% of the principal. The principal is \$300 million, so each payment is \$12 million. The first payment is in October 2011, so in 2011, only one payment of \$12 million would be made. [...] So total capital cost would be \$8.4 million + \$12 million = \$20.4 million. 

[...]

Alternatively, [...] the amortization of the offering costs is \$8.4 million over 8 years, so for the first year, that's \$1.05 million. Therefore, [...] the interest paid would be \$12 million (coupon) + \$1.05 million (amortization) = \$13.05 million. The offering costs are \$8.4 million. So total capital cost would be \$8.4 million + \$13.05 million = \$21.45 million.

Alternatively, [...] the offering costs are \$8.4 million. The interest expense for the first year (2011) would be the interest paid on the senior notes 

[...] 

 \textbf{(Reach maximum token limits)}
\tcbline
~
\\*[0.5mm]
\textbf{Grounding Behavior:} 
\\*[3mm]
\textbf{Explanation of Grounding Behavior:}
\begin{enumerate}[leftmargin=0.5cm]
    \item \textbf{"Looking through the financial statements, in Note 7 under Long-Term Debt, there's a section about the 8.0\% Senior Notes due 2019."}  
        \begin{itemize}
            \item \textit{Explicit Grounding:} The model directly references the financial statements and Note 7 to locate information about the senior notes.
        \end{itemize}
    
    \item \textbf{"Under 'Proceeds from debt offering,' it says they received net proceeds of \$291.6 million after deducting a discount of \$7.5 million and offering expenses of \$0.9 million."}  
        \begin{itemize}
            \item \textit{Explicit Grounding:} The model extracts specific details from the document about the net proceeds, discount, and offering expenses.
        \end{itemize}
    
    \item \textbf{"The interest expense for the six months ended June 30, 2011, is listed as \$8,124 thousand, but that's semi-annual."}  
        \begin{itemize}
            \item \textit{Explicit Grounding:} The model references the financial statements to find the interest expense for the six months ended June 30, 2011.
        \end{itemize}
    
    \item \textbf{"Wait, the notes mention that interest is payable semi-annually starting October 15, 2011."}  
        \begin{itemize}
            \item \textit{Explicit Grounding:} The model refers back to the notes to verify the timing of interest payments.
        \end{itemize}
    
    \item \textbf{"The interest expense for the three months ended June 30, 2011, under 'Interest expense' is \$8,124 thousand."}  
        \begin{itemize}
            \item \textit{Explicit Grounding:} The model again references the financial statements to confirm the interest expense for the three months ended June 30, 2011.
        \end{itemize}
    
    \item \textbf{"Wait, the notes were issued on March 22, 2011, and the first interest payment is on October 15, 2011."}  
        \begin{itemize}
            \item \textit{Explicit Grounding:} The model returns to the document to verify the issuance date and the first interest payment date.
        \end{itemize}
    
    \item \textbf{"Let's check the Notes again. In Note 7, under the senior notes due 2019, it says the net proceeds were \$291.6 million, after deducting a discount of \$7.5 million and offering expenses of \$0.9 million."}  
        \begin{itemize}
            \item \textit{Explicit Grounding:} The model revisits Note 7 to confirm the details about the net proceeds, discount, and offering expenses.
        \end{itemize}
    
    \item \textbf{"Let's look at the interest expense line in the Statements of Operations."}  
        \begin{itemize}
            \item \textit{Explicit Grounding:} The model refers to the Statements of Operations to verify the interest expense.
        \end{itemize}
    
    \item \textbf{"In the Condensed Consolidated Statements of Operations for the six months ended June 30, 2011, the interest expense is \$13,283 thousand."}  
        \begin{itemize}
            \item \textit{Explicit Grounding:} The model extracts the interest expense figure from the financial statements.
        \end{itemize}
    
    \item \textbf{"In Note 7, under 'Interest expense,' it mentions that the amortization of the discount and deferred loan costs is included in interest expense."}  
        \begin{itemize}
            \item \textit{Explicit Grounding:} The model refers back to Note 7 to confirm how the amortization of discount and expenses is treated.
        \end{itemize}
    
    \item \textbf{"Let's check the Notes again. In Note 7, the senior notes' discount and expenses are amortized over the life of the notes."}  
        \begin{itemize}
            \item \textit{Explicit Grounding:} The model revisits Note 7 to verify the amortization period for the discount and expenses.
        \end{itemize}
    
    \item \textbf{"Let's check the interest expense in the Statements of Operations. For the six months ended June 30, 2011, the interest expense is \$13,283 thousand."}  
        \begin{itemize}
            \item \textit{Explicit Grounding:} The model refers to the Statements of Operations to confirm the interest expense figure.
        \end{itemize}
\end{enumerate}

Each of these steps involves explicit grounding behavior, where the model directly references or returns to the original document to verify or extract information.

\tcbline
\textbf{{\modelname}-14B:} 
\\*[3mm]
\textbf{Response (5304 tokens):} 
\\*[3mm]
\colorbox{case1_color}{<think>}

Alright, so I need to figure out the total capital cost for EV Energy Partners, L.P. in the first year, combining both the offering cost of the senior notes and the interest payment. Let me go through the provided financial statements step by step to find the necessary information. 

First, I'll look at the Notes to the Condensed Consolidated Financial Statements. In Note 7, under "8.0\% Senior Notes due 2019," [...] The discount was \$7.5 million, and the offering expenses were \$0.9 million. So, the total offering costs here would be \$7.5 million (discount) + \$0.9 million (expenses) = \$8.4 million.

Next, I need to find the interest payment for the first year. 

[...] 

The notes were issued in March 2011, so the full year's interest would be 8\% of \$300 million, which is \$24 million. But since the interest is semi-annual and the first payment was in October 2011, the interest for the first year (2011) might only be half of that, so \$12 million. \textcolor{red}{Wait, no, that's not correct because the interest is calculated on the outstanding principal for each period. Let me think again.}

[...] 

The senior notes have a fixed rate of 8.0\%. The interest expense for the senior notes would be 8\% of \$300 million, which is \$24 million annually [...]. I think we should consider the full year's interest, which would be \$24 million. 

[...]

I'm going to go with the full year's interest, which is \$24 million, plus the offering costs of \$8.4 million, totaling \$32.4 million.

\colorbox{case1_color}{</think>}

The total capital cost includes both the offering costs and the interest payment for the first year. The offering costs for the senior notes were \$8.4 million (discount of \$7.5 million and expenses of \$0.9 million). The interest payment for the first year (2011) is the full annual interest of 8\% on the \$300 million senior notes, which is \$24 million. Adding these together gives a total capital cost of \$32.4 million.

Therefore, the answer is \boxed{\$32.4} million.
\tcbline
~
\\*[0.5mm]
\textbf{Grounding Behavior:} 
\\*[3mm]
\textbf{Explanation of Grounding Behavior:}

\begin{enumerate}[leftmargin=0.5cm]
    \item \textbf{Explicit Grounding}: The model repeatedly references specific sections of the document, such as "Note 7," "Condensed Consolidated Financial Statements," and "Condensed Consolidated Statements of Operations," to extract information about the senior notes, offering costs, and interest expenses. For example:
    \begin{itemize}
        \item "First, I'll look at the Notes to the Condensed Consolidated Financial Statements. In Note 7, under `8.0\% Senior Notes due 2019,' it mentions that on March 22, 2011, they issued \$300 million in aggregate principal amount of senior notes."
        \item "Looking at the financial statements, in the Condensed Consolidated Statements of Operations, under `Interest expense,' for the six months ended June 30, 2011, it's \$13,283 thousand."
    \end{itemize}
    
    \item \textbf{Grounding Verification}: The model frequently double-checks its understanding by returning to the document to verify details, such as the timing of interest payments, the breakdown of interest expenses, and the calculation of offering costs. For example:
    \begin{itemize}
        \item "Wait, in the financial statements, for the six months ended June 30, 2011, the interest expense is \$13,283 thousand. Let me check if that includes the senior notes."
        \item "Wait, in Note 7, it says that the net proceeds from the senior notes were \$291.6 million, after deducting the discount of \$7.5 million and offering expenses of \$0.9 million."
    \end{itemize}
    
    \item \textbf{Repeated Verifications}: The model revisits the same sections multiple times to confirm calculations or clarify ambiguities, such as the interest payment for the first year and the breakdown of offering costs. For example:
    \begin{itemize}
        \item "Wait, let me go back to the question: `What is the total capital cost, combining both the offering cost of the senior notes and interest payment in the first year, in million dollars?"'
        \item "Wait, in the financial statements, the interest expense for the six months ended June 30, 2011, is \$13,283 thousand. Let me check if that includes the senior notes."
    \end{itemize}
\end{enumerate}
\end{tcolorbox}

\begin{tcolorbox}[
  title=DocMath complong-testmini-265,
  breakable,   
  fonttitle=\bfseries,
  enhanced,                        
  colback=case2_color!10,           
  colbacktitle=case2_color,         
  coltitle=black,                 
  colframe=case2_color!80!black,    
  coltext=black,                  
  boxrule=0.5pt,
  arc=2mm
]
\small
\textbf{Ground Truth:} 980000.0

\textbf{Prompt:} Please read the following text and answer the question below.

\colorbox{case2_color}{<text>}

{\centering ...\textit{329Docs}... \par}

330: In connection with our July 2021 acquisition of 100\% of the equity of TopPop, on July 26, 2021, we issued to the sellers promissory notes in the aggregate principal amount of \$4,900,000 (the "TopPop Notes"). The TopPop Notes bear interest at the rate of 10\% per annum, matured on July 26, 2022 and are secured by all of the outstanding membership interest in TopPop. Under the terms of the TopPop Notes, we have a five-day grace period to July 31, 2022 before an event of default under the TopPop Notes occurs. Upon an event of default under the TopPop Notes, the holders of such TopPop Notes may exercise all rights and remedies available under the terms of the TopPop Notes or applicable laws, including to foreclose on certain collateral consisting of the membership interests of TopPop. On July 26, 2022, the total principal amount outstanding under the TopPop Notes was \$4,900,000, exclusive of accrued and unpaid interest.

 331: We are currently in discussions with holders of the TopPop Notes regarding possible solutions for the payment of the TopPop Notes, including the possible extension of the maturity date of the TopPop Notes for an additional year. There can be no assurance that our discussions will be successful and if we are not successful in finding an acceptable resolution to the existing default or the impending event of default, the noteholders will be able to seek judgement for the full amount due and may seek to foreclose on our assets. If this occurs, any such remedy will have a material adverse effect on our business, results of operations and financial condition and is likely to negatively impact the price of our common stock. Holders of approximately \$3.55 million of these notes have agreed to extend the term for 30 days and have indicated that they will not seek cash settlement prior to August 2023.
 
{\centering ...\textit{30Docs}... \par}

\colorbox{case2_color}{</text>}

How much total interest will the company have to pay under the new agreement until August 2023 in dollars if the company managed to extend the "TopPop Notes" maturity date for one more year?

Format your response as follows: "Therefore, the answer is (insert answer here)".

\tcbline
~
\\*[1mm]
\textbf{{\modelname}-14B:} 
\\*[3mm]
\textbf{Response (2192 tokens):} 
\\*[3mm]
\colorbox{case2_color}{<think>}

Okay, so I need to figure out how much total interest the company will have to pay under the new agreement until August 2023 if they extend the "TopPop Notes" maturity date for one more year. Let me start by understanding the problem.

From the text, I see that the TopPop Notes were issued on July 26, 2021, with a principal amount of \$4,900,000. They bear an interest rate of 10\% per annum. The original maturity was on July 26, 2022, but there's a five-day grace period until July 31, 2022, before a default occurs. However, the company is in discussions to extend the maturity date by an additional year, so the new maturity would be around August 2023.

I need to calculate the interest for this extended period. The original term was from July 26, 2021, to July 26, 2022, which is one year. If they extend it by one more year, the new maturity would be July 26, 2023, but the user mentioned until August 2023, so I'll consider the period up to August 2023.

First, I should determine the exact period for which the interest is to be calculated. The original notes were from July 26, 2021, to July 26, 2022. If extended for another year, the new maturity is July 26, 2023, but the user specifies until August 2023, which is a bit beyond that. However, I think the interest is calculated up to the maturity date, so maybe just until July 26, 2023, unless the extension is until August, in which case I need to adjust.

But the user says the company managed to extend the maturity date for one more year, so the total period would be two years from the original issue date. Let me check the details again.

Wait, the original notes matured on July 26, 2022, and the company is discussing an extension for an additional year. So the new maturity would be July 26, 2023, making the total term two years. But the user mentions until August 2023, so perhaps the extension is until August 31, 2023, or just an extra year beyond the original 2022 date.

I think the key is to calculate the interest for the two-year period from July 26, 2021, to July 26, 2023, and then maybe add a bit more if the extension is beyond that. But let's focus on the period until August 2023.

The interest rate is 10\% per annum, and the notes are simple interest, I believe, unless stated otherwise. The interest is calculated on the principal amount, which is \$4,900,000.

So, the interest for one year would be \$4,900,000 * 10\% = \$490,000. If they extend it for another year, that's another \$490,000, making the total interest \$980,000.

But wait, the original notes had a maturity in 2022, and the extension is for one more year, so from 2022 to 2023. Let me check if the interest is only for the extended year or for the entire two years.

Looking at the text, it says the TopPop Notes bear interest at 10\% per annum and mature on July 26, 2022. The extension is for an additional year, so the interest would be for the period from July 26, 2022, to July 26, 2023, which is one year. Therefore, the interest for the extended period would be \$490,000.

But the user is asking for the total interest until August 2023. So if the extension is until August 2023, that's a bit more than a year. However, typically, interest is calculated on an annual basis, and if the extension is for a full year, the interest would be \$490,000. If the extension is only until August 2023, that's about 13 months, so the interest would be \$4,900,000 * 10\% * (13/12) $\approx$ \$508,333.33, but I'm not sure if the extension is for a full year or just until August.

Wait, in the text under Note 7, it says: "The Company has agreed to keep this agreement in place and for the three months ended March 31, 2022, the Company has recorded consulting expense of \$45,000." That's about the leasehold improvements, not the TopPop Notes.

Looking again at Note 14, which is under the leasehold improvements section, I don't see any specific info on the TopPop Notes beyond their principal and interest rate. So, I'll assume that the extension is for an additional year beyond the original 2022 maturity, making the total term two years.

Therefore, the total interest would be \$490,000 for the first year and another \$490,000 for the second year, totaling \$980,000.

Wait, but in Note 7, under the leasehold improvements, it says: "The Company has agreed to keep this agreement in place and for the three months ended March 31, 2022, the Company has recorded consulting expense of \$45,000." No, that's a different note. I'm getting confused.

Wait, let me go back to the relevant sections. In Note 7 (which is actually under Item 7, line 15), it says: "The Promissory Notes bear interest at the rate of 10\% per annum and mature on July 26, 2022." Then, in the Recent Developments section (Item 2, line 25), it mentions that the TopPop Notes have a maturity on July 26, 2022, and that the company is in discussions to possibly extend the maturity date for an additional year.

So, if the maturity is extended to August 2023, that's an additional year beyond 2022. Therefore, the interest for the original year (2021-2022) was \$490,000. Now, with the extension, they would have to pay interest for another year, so another \$490,000, totaling \$980,000.

But wait, in the text under Item 2, line 25, it says: "the Company is currently in discussions with holders regarding possible solutions for the payment of the Promissory Notes, including the possible extension for an additional year." So, the extension is for an additional year beyond the original 2022 maturity, making the new maturity July 26, 2023.

So, the interest would be for two years: from July 26, 2021, to July 26, 2023. Therefore, total interest would be \$4,900,000 * 10\% * 2 = \$980,000.

But let me double-check. The original maturity was July 26, 2022, and the extension is for an additional year, so the notes would now mature on July 26, 2023. Therefore, the interest would be for the period from July 26, 2021, to July 26, 2023, which is exactly two years.

So, the total interest would be \$4,900,000 * 10\% * 2 = \$980,000.

Wait, but in the text under Item 15, line 15, it says: "the Company was late in filing its Annual Report on Form 10-K for the year ended December 31, 2021, which was filed on June 15, 2022. During the first quarter ended March 31, 2022, there was no business activity or transactions." That doesn't affect the interest calculation.

Another point: the notes are secured by the membership interests in TopPop, but the interest calculation is straightforward. It's simple interest unless compounded, which isn't mentioned here. The text doesn't specify compounding, so I'll assume it's simple interest.

So, for each year, the interest is 10\% of the principal. Since the principal remains \$4,900,000, and the period is two years, the total interest is \$490,000 per year * 2 = \$980,000.

Wait, but in the recent developments (Item 2, line 25), it mentions that the TopPop Notes have a maturity on July 26, 2022, and the company is discussing an extension for an additional year. So, the extension would be until July 26, 2023, making it a two-year period.

Therefore, the total interest the company would have to pay until August 2023 under the new agreement is \$980,000.

\colorbox{case2_color}{</think>}

The TopPop Notes have a principal amount of \$4,900,000 and an annual interest rate of 10\%. If the maturity date is extended for one additional year until August 2023, the total interest for the two-year period would be \$490,000 per year, totaling \$980,000.

Therefore, the answer is \boxed{\$980,000}.

\tcbline
~
\\*[0.5mm]
\textbf{Grounding Behavior:} 
\\*[3mm]
\textbf{Explanation of Grounding Behavior:}

\begin{enumerate}[leftmargin=0.5cm]
    \item "From the text, I see that the TopPop Notes were issued on July 26, 2021, with a principal amount of \$4,900,000."
    
    \item "Looking at the text, it says the TopPop Notes bear interest at 10\% per annum and mature on July 26, 2022."
    
    \item "Wait, in the text under Note 7, it says: 'The Company has agreed to keep this agreement in place and for the three months ended March 31, 2022, the Company has recorded consulting expense of \$45,000.'"
    
    \item "Looking again at Note 14, which is under the leasehold improvements section, I don't see any specific info on the TopPop Notes beyond their principal and interest rate."
    
    \item "Wait, let me go back to the relevant sections. In Note 7 (which is actually under Item 7, line 15), it says: 'The Promissory Notes bear interest at the rate of 10\% per annum and mature on July 26, 2022.'"
\end{enumerate}
\tcbline
~
\\*[0.5mm]
\textbf{Backtracking Behavior:} 
\\*[0.3mm]
\begin{enumerate}[leftmargin=0.5cm]
    \item \textbf{First Backtracking}: The model initially considers the period up to August 2023 but then questions whether the interest is calculated up to the maturity date (July 26, 2023) or beyond. It revises its approach to focus on the two-year period from July 26, 2021, to July 26, 2023.
    
    \item \textbf{Second Backtracking}: The model initially calculates the interest for two years but then questions whether the interest is only for the extended year (2022-2023) or for the entire two years. It revises its approach to consider the interest for the extended year only.
    
    \item \textbf{Third Backtracking}: The model considers the possibility of the extension being until August 2023 (13 months) and calculates the interest accordingly. However, it then questions whether the extension is for a full year or just until August and revises its approach to assume a full year.
    
    \item \textbf{Fourth Backtracking}: The model mistakenly references Note 7 (leasehold improvements) instead of the relevant section for the TopPop Notes. It corrects itself by going back to the relevant sections (Note 14 and Item 2).
    
    \item \textbf{Fifth Backtracking}: The model initially calculates the total interest as \$980,000 but then double-checks the period and confirms that the interest is for two years (July 26, 2021, to July 26, 2023).
    
    \item \textbf{Sixth Backtracking}: The model considers the possibility of compounded interest but revises its approach to assume simple interest since compounding is not mentioned in the text.
\end{enumerate}
\tcbline
~
\\*[0.5mm]
\textbf{Verification Behavior:} 
\\*[3mm]
The chain-of-reasoning provided contains several instances where the model checks and verifies intermediate results or reasoning steps. Here are the key instances of verification behavior:
\begin{enumerate}[leftmargin=0.5cm]
    \item \textbf{Verification of the period for interest calculation}:
    \begin{itemize}
        \item The model checks the exact period for interest calculation, considering both original and extended maturity dates.
        \item Example statements: 
        \begin{quote}
            "Let me check the details again" \\
            "Wait, the original notes matured on July 26, 2022, and the company is discussing an extension for an additional year."
        \end{quote}
    \end{itemize}
    
    \item \textbf{Verification of the interest calculation}:
    \begin{itemize}
        \item The model repeatedly verifies the interest calculation for the extended period.
        \item Example statements:
        \begin{quote}
            "So, the interest for one year would be \$4,900,000 * 10\% = \$490,000. If they extend it for another year, that's another \$490,000, making the total interest \$980,000." \\
            "Wait, but in the text under Item 2, line 25, it says: 'the Company is currently in discussions with holders regarding possible solutions for the payment of the Promissory Notes, including the possible extension for an additional year.'"
        \end{quote}
    \end{itemize}
    
    \item \textbf{Verification of the extension period}:
    \begin{itemize}
        \item The model examines whether the extension is for a full year or partial period.
        \item Example statement:
        \begin{quote}
            "But the user is asking for the total interest until August 2023. So if the extension is until August 2023, that's a bit more than a year."
        \end{quote}
    \end{itemize}
    
    \item \textbf{Verification of the interest type}:
    \begin{itemize}
        \item The model confirms whether simple or compound interest applies.
        \item Example statement:
        \begin{quote}
            "The text doesn't specify compounding, so I'll assume it's simple interest."
        \end{quote}
    \end{itemize}
    
    \item \textbf{Final verification of the total interest}:
    \begin{itemize}
        \item The model double-checks the complete interest calculation.
        \item Example statements:
        \begin{quote}
            "So, the total interest would be \$4,900,000 * 10\% * 2 = \$980,000." \\
            "Wait, but in the recent developments (Item 2, line 25), it mentions that the TopPop Notes have a maturity on July 26, 2022, and the company is discussing an extension for an additional year."
        \end{quote}
    \end{itemize}
\end{enumerate}
\tcbline
~
\\*[0.5mm]
\textbf{Subgoal Setting Behavior:} 
\\*[0.3mm]
\begin{enumerate}[leftmargin=0.5cm]
    \item \textbf{Understand the problem}: The model starts by identifying the need to calculate the total interest under the new agreement.
    
    \item \textbf{Determine the period for interest calculation}: The model breaks down the task by figuring out the exact period for which the interest needs to be calculated, considering the original and extended maturity dates.
    
    \item \textbf{Clarify the extension details}: The model attempts to clarify whether the extension is for a full year or until August 2023, which affects the interest calculation.
    
    \item \textbf{Calculate interest for the extended period}: The model calculates the interest for the extended period, considering whether it is a full year or a partial year.
    
    \item \textbf{Verify assumptions}: The model checks the text to confirm assumptions about the interest rate, compounding, and the exact period of the extension.
    
    \item \textbf{Finalize the total interest}: The model concludes by calculating the total interest for the two-year period based on the verified assumptions.
\end{enumerate}

\end{tcolorbox}


\bibliographystyle{plainnat}
\bibliography{neurips_2023}

\end{document}